\documentclass[final]{colt2022} 

\usepackage{macros-colt}
\usepackage{booktabs}
\usepackage{floatrow}
\newfloatcommand{capbtabbox}{table}[][\linewidth]

\let\todo\undefined

\usepackage{enumitem}

\title[Knowledge Distillation: Bad Models Can Be Good Role Models]{Knowledge Distillation: Bad Models Can Be Good Role Models}
\usepackage{times}

\coltauthor{%
 \Name{Gal Kaplun} \Email{galkaplun@g.harvard.edu}\\
 \addr Harvard University \& Mobileye 
 \AND
 \Name{Eran Malach} \Email{eran.malach@mail.huji.ac.il}\\
 \addr Hebrew University \& Mobileye 
 \AND
 \Name{Preetum Nakkiran} \Email{preetum@ucsd.edu}\\
 \addr University of California San Diego
 \AND
 \Name{Shai Shalev-Shwartz} \Email{shais@cs.huji.ac.il}\\
 \addr Hebrew University \& Mobileye 
}

\begin{document}

\maketitle

\begin{abstract}%
Large neural networks trained in the overparameterized regime are able to fit noise to zero train error. Recent work \citep{nakkiran2020distributional} has empirically observed that such networks behave as ``conditional samplers'' from the noisy distribution. That is, they replicate the noise in the train data to unseen examples.
We give a theoretical framework for studying this conditional sampling behavior in the context of learning theory. We relate the notion of such samplers to knowledge distillation, where a student network imitates the outputs of a teacher on unlabeled data. We show that samplers, while being bad classifiers, can be good teachers. Concretely, we prove that distillation from samplers is guaranteed to produce a student which approximates the Bayes optimal classifier. Finally, we show that some common learning algorithms (e.g., Nearest-Neighbours and Kernel Machines) can generate samplers when applied in the overparameterized regime.
\end{abstract}

\begin{keywords}%
  Knowledge Distillation, Deep Learning, Label Noise
\end{keywords}
\section{Introduction}

Recently, the field of supervised learning has witnessed the success of \emph{overparameterized} methods:
models, like deep neural networks, which are large enough to fit their train set
but still achieve good test performance.
A core theoretical concern is to understand why such models are able to fit even noisy training data without catastrophically overfitting \citep{zhang2017} despite no explicit regularization.
The seminal work of \cite{bartlett2020benign}
proposed the theoretical framework of \emph{benign overfitting}
to capture this empirical behavior.
Briefly, benign overfitting studies \emph{statistically consistent} methods---where models approach the Bayes optimal function, even in presence of noise.

\begin{figure}[ht!]
    \centering
    \includegraphics[width=0.31\textwidth]{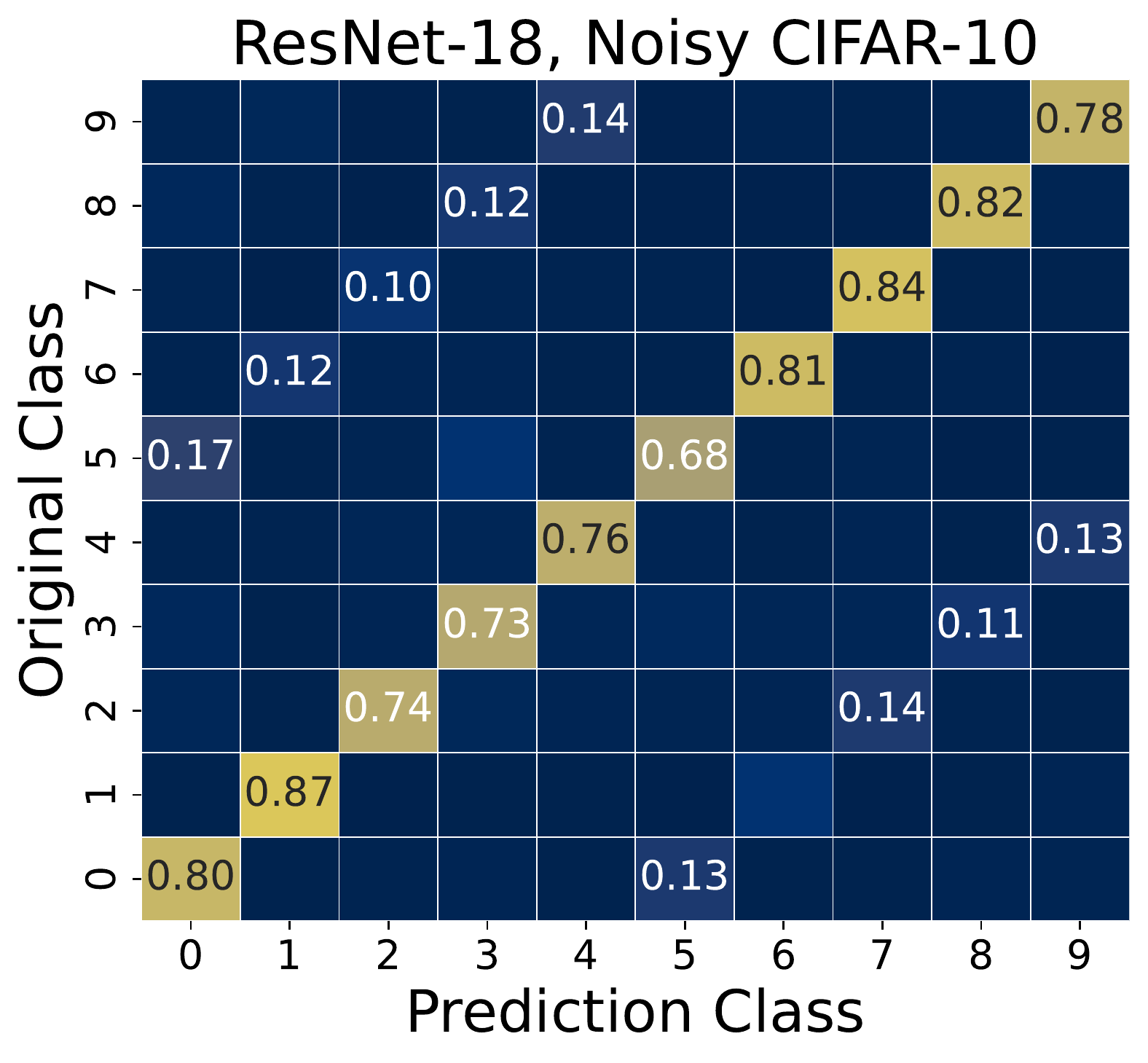} \hfil 
    \includegraphics[width=0.34\textwidth]{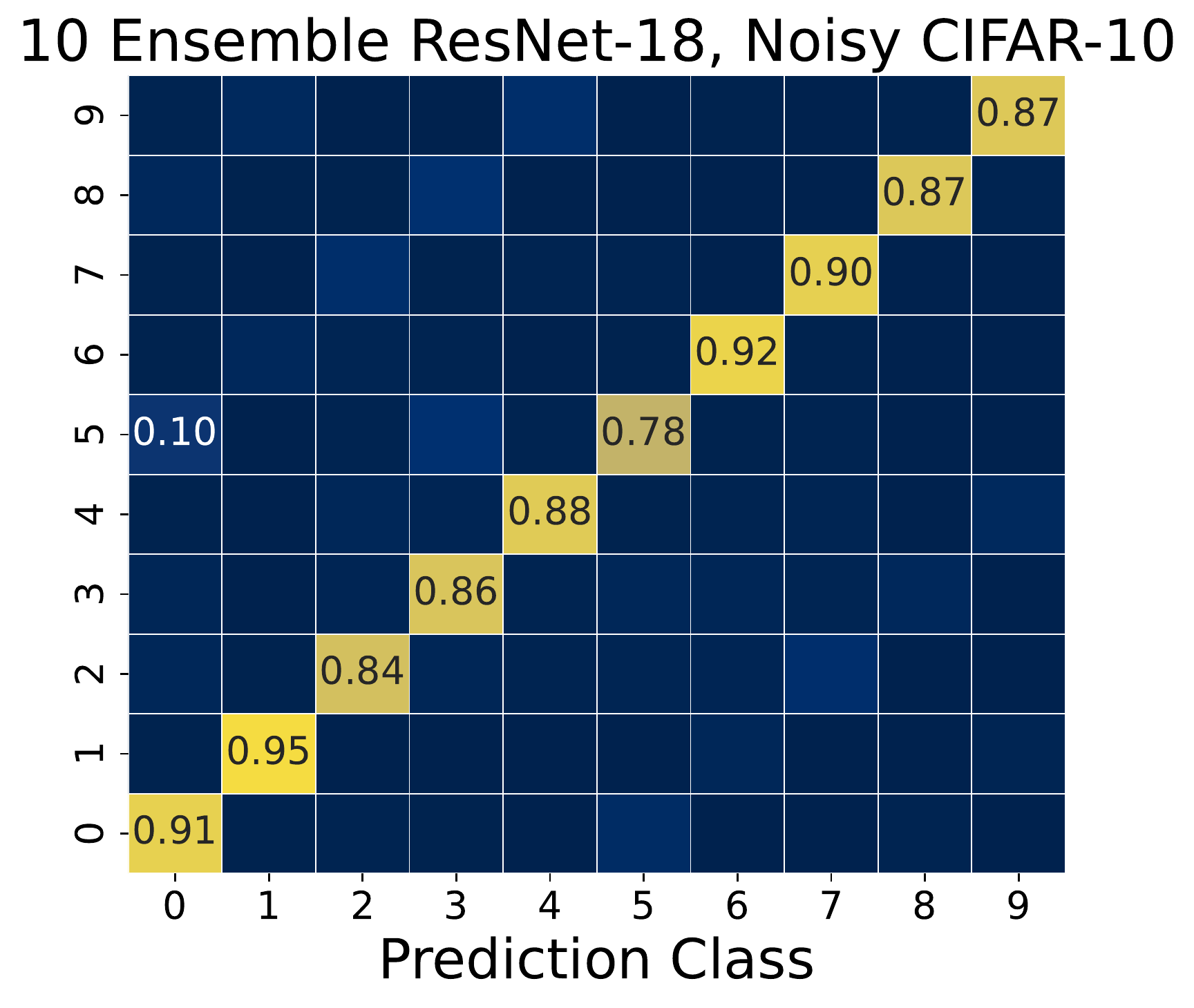} \hfil 
    \includegraphics[width=0.32\textwidth]{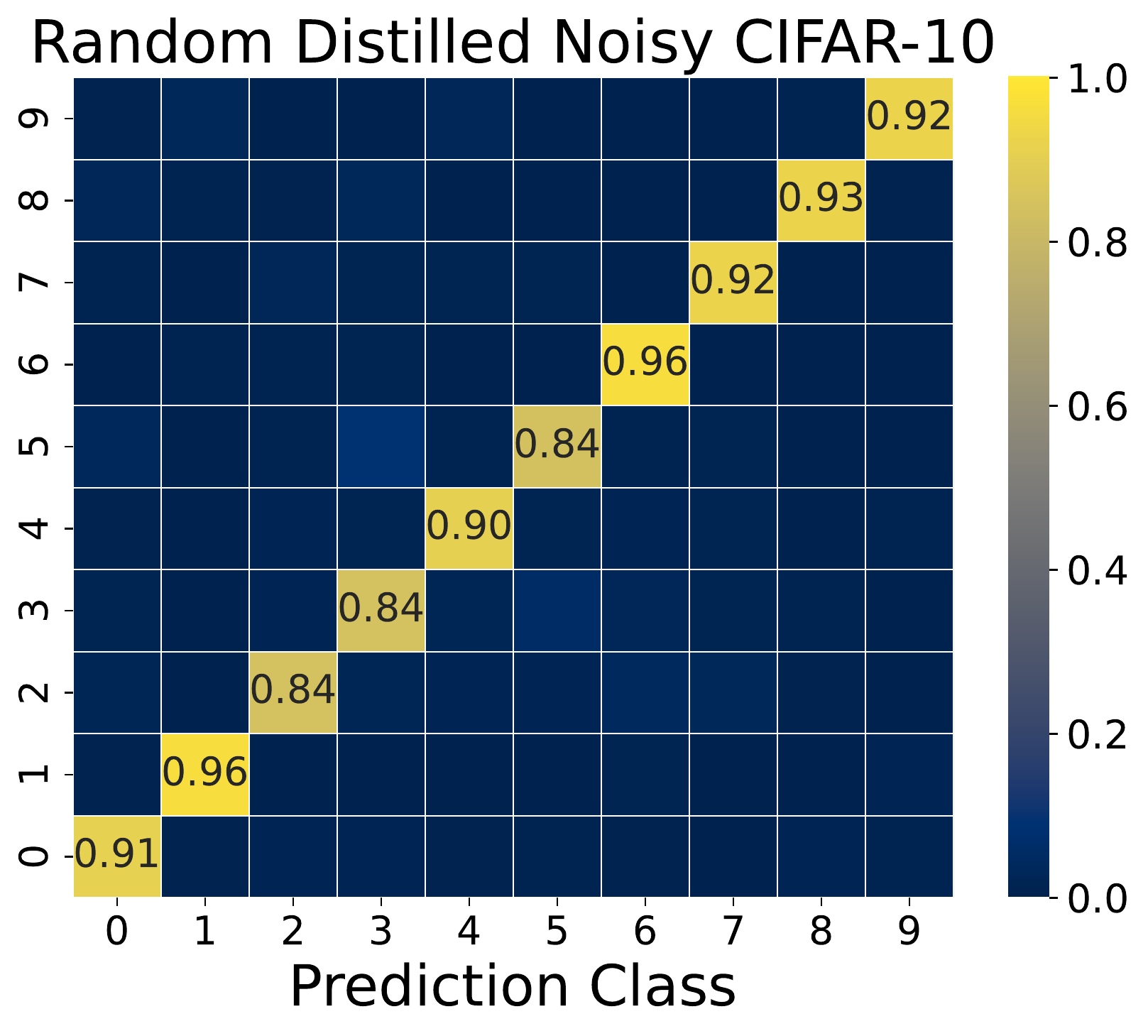}\hfil
    \caption{\textbf{From Samplers to Good Models.}  Class confusion matrices for ResNet-18 trained on CIFAR-10 with 20\% fixed and class dependent label noise (e.g., 20\% of cats are labeled as dogs). \emph{Left.} A standalone ResNet-18 trained on this data \emph{replicates} the noise to the test distribution. \emph{Middle.} When ensembling 10 such models, the noise virtually disappears, however incurring a high price at inference. \emph{Right.} Distilling a single model via unlabeled examples from the CIFAR-5m dataset using a randomly selected teacher from a pool of 10 such teachers  for each example eliminates the noise as well as the inference cost.\vspace{-0.6cm}
    }
    \label{fig:intro}
\end{figure}

However, recent empirical work shows that 
when training deep neural networks on noisy data,
\emph{overfitting is neither catastrophic nor benign} \citep{nakkiran2020distributional}.
Specifically, they propose that overfitting leads 
not to a good classifier, but to a good \emph{conditional sampler}.
For example, suppose we train a model on a set of images sampled from some distribution $\cd$, where $20\%$ of the images of cats are wrongly labeled as dogs.
We now train an overparameterized network to fit samples from $\cD$.
Note that for the distribution $\cD$, the Bayes-optimal classifier, namely $f^*_\cd(\x) := \argmax_y \P_\cd(y | \x)$, returns the ``correct'' class of every image.
We can hope that if overfitting is truly ``benign'' in the sense of \citet{bartlett2020benign}, 
the overparameterized model will be close to this optimal $f^*_\cd$.
However, this is not what occurs in practice: 
as \citet{nakkiran2020distributional} point out,
the trained model $f$ reproduces noise in the training set at test time, labeling up to 20\% of the cats in the \emph{test} data as dogs.
In a sense, the trained model $f$ behaves as a conditional sample: $f(\x) \sim \P_\cd(y|\x)$ (see the leftmost confusion matrix in Figure \ref{fig:intro}).

The above example indicates that thinking about classifiers in the overparameterized regime as approximating the Bayes optimal predictor might be misleading.
Therefore, it is essential to 
develop the appropriate theoretical framework for describing the behavior of \emph{samplers} from the conditional distribution.
While the learning-theoretic aspects of supervised classification 
are well-studied, the theory of supervised \emph{conditional sampling}
has, to the best of our knowledge, not been systematically explored.
In this work, we take the first steps towards addressing this gap.
We initiate the study of conditional sampling as a learning problem,
and explore its relations to other kinds of learning.

Interestingly, we relate the notion of samplers to  knowledge distillation methods. In short, knowledge distillation is the process of training a teacher network on a small labeled dataset and using its predictions to label a large unlabeled dataset, on which a student network is trained to imitate the output of the teacher. We find that taking an ensemble of samplers as a teacher for knowledge distillation produces a student network with minimal error with respect to the Bayes optimal classifier. Finally, our theory leads to a new algorithm for knowledge distillation, where we randomly choose a teacher from a fixed pool, to label each example, which accelerates the training process in practice. We show that this new algorithm is guaranteed to find a student with low error. 

\subsection{Our Contributions}

\paragraph{Learning Theory of Conditional Samplers. (Section~\ref{sec:setting})}
We initiate the study of \emph{conditional sampling} in the context of computational learning theory.
We formally introduce the problem of conditional-sampling, and define the sample complexity
of learning a sampler.
We then present positive and negative results in this new setting.
For example, we show that there exist distributions where
\emph{sampling} is much easier than \emph{classification},
requiring far fewer samples.
However, if we allow polynomial blowup in sample-size and runtime, a sampler can be ``boosted'' to a good classifier, showing that if finding a classifier is computationally hard then sampling is also hard.



\paragraph{Theory of Knowledge Distillation. (Section~\ref{sec:distillation})}
One way to boost a sampler into a good classifier is to run an ensemble of samplers at inference time (see middle panel of Figure~\ref{fig:intro}), which is costly. We show that by performing distillation from an ensemble of teachers, it is possible to find a student with low error w.r.t. the Bayes optimal. This shows that teacher networks in ensemble-distillation need not be good classifiers, but just good samplers.
Finally we propose a new algorithm for distillation, where each example is labeled by a random teacher from a fixed pool (see right panel of Figure~\ref{fig:intro}). We study quantitative bounds on the sample complexity of teaching and learning in our setting, for both ensemble-distillation and distillation from a random teacher.

\paragraph{Theory of Sampler Algorithms. (Section~\ref{sec:algorithms})}
We show that several classical learning algorithms provably produce good conditional samplers,
and analyze their sample complexity in terms of standard problem parameters
(e.g., distributional smoothness). Specifically we show that the 1-Nearest-Neighbour algorithm is a sampler, and extend our analysis to k-Nearest-Neighbour as well. Furthermore, we show that under some distributional assumptions, Lipschitz classes such as linear methods, kernels and neural networks may also behave like samplers.
\subsection{Related Work}
\paragraph{Knowledge Distillation}
Knowledge Distillation for deep learning was proposed in \citet{hinton2015distilling}. Since then, a large body of work showed its practical benefits for various machine learning tasks \citep{pseudolabels2019,metapsedulabels2020, yalniz2019billionscale, simclr2, furlanello2018born, nlp2}. Nonetheless, from a theoretical perspective, understanding why and when distillation works remains a mystery. \cite{hinton2015distilling} attribute the success of distillation to the fact that the soft labels of the teacher passing additional information on the input. In a recent paper, \cite{menon2020} claim that when the teacher approximates the Bayes class-probabilities, distillation is possible. Our results show that even when the teachers are far from the Bayes optimal prediction (i.e., when teachers are noisy samplers from the distribution), we can find a student with low levels of noise through distillation. Another work by \cite{wei2020theoretical} shows that, assuming the data distribution has good continuity within each class, self-distillation is possible. However, it is not clear when such assumption is actually satisfied. The work of \cite{lopez2015unifying} relates the notion of privileged information to knowledge distillation. A work by \cite{frei2021self} shows that self-training can boost weak learners, when the target is a linear classifier over a mixture model distribution. Other works \citep{mobahi2020self, zhang2022does, dong2019distillation} study distillation as a regularization method, forcing the student to learn under a ``smoother'' loss landscape.

\paragraph{Learning with Noise} Learning under corrupted labels is a well-studied research area in the literature (e.g., \cite{angluin1988learning, frenay2013classification}). A notable example is the work of \cite{blum2003noise, kearns1998efficient}, showing how statistical query algorithms can be leveraged to learn under noisy labels. A growing number of works study the effects of noise on deep learning, as well as methods to learn under label noise (see \cite{song2020learning} for a survey). However, most of these works do not leverage distillation for label noise robustness, as we suggest in our work. More related to our work is a work by \cite{haase2021iterative}, showing that an iterative process of training and re-labeling can combat label noise. However, this work focuses on empirical study.

\section{Sampling as a Learning Problem}
\label{sec:setting}
Let $\cx$ be the input space and $\cy = \{\pm 1\}$ be the label space (for simplicity, under appropriate assumptions we can extend most of our results to multiclass). A hypothesis class $\ch$ is some class of functions from $\cx$ to $\cy$. A learning algorithm $\ca$ takes a sequence of $m$ samples $S \in (\cx \times \cy)^m$ and outputs some hypothesis $h : \cx \to \cy$. We denote by $\ca(S)$ the hypothesis that $\ca$ outputs when observing the sample $S$.
For some distribution $\cd$ over $\cx \times \cy$, we denote its $\cx$ marginal as $\cd_\cx$ and denote the Bayes optimal classifier by $
f^*_\cd(\x) := \arg \max_{y \in \cy} \prob{\cd}{y | \x}$\footnote{We assume that ties are broken arbitrarily}.

When $\cd$ is a distribution where the label is not a deterministic function of the input, we think of $\cd$ as a noisy version of some clean distribution $\cd^*$, where each input is correctly labeled. Naturally, we assume that the probability of seeing the right label is greater than seeing a wrong one. In other words, $\cd^*$ has the same marginal distribution over $\cx$ (i.e., $\cd_\cx = \cd^*_\cx$), and is labeled by the Bayes optimal classifier of $\cd$. That is, sampling $(\x,y) \sim \cd^*$ is given by $\x \sim \cd_\cx$ and $y = f^*_\cd(\x)$.

We denote the ``noise'' of the distribution $\cd$ by $\eta(\cd) := \prob{(\x,y) \sim \cd}{y \ne f^*_\cd(\x)}$. We also consider the margin of the distribution, which defines the difference in probability between correct and wrong label for each example. Namely, for some $\delta \ge 0$, let $\gamma_\delta(\cd)$ be the supremum over $\gamma >0$ s.t.,
\[
\P_{\x \sim \cd_\cx}\left[\P_{\cd}(f^*_\cd(\x)|\x) < \max_{y \ne f^*_\cd(\x)} \P_{\cd}(y|\x) + \gamma \right] \le \delta
\]
Specifically, we denote $\gamma(\cd) := \gamma_0(\cd)$. 
We typically assume that the input distribution has a strictly positive margin $\gamma(\cd) > 0$, so for every example the probability to see the correct label is greater by $\gamma$ than the probability to see a wrong label.

Our objective in this setting is to approximate the Bayes optimal classifier of $\cd$. In other words, we want to minimize the $0$-$1$ loss on the clean distribution $\cd^*$:
\begin{equation}
\label{eq:loss}
L_{\cd^*}(h) := \prob{(\x,y) \sim \cd}{h(\x) \ne f^*_\cd(\x)} = \prob{(\x,y) \sim \cd^*}{h(\x) \ne y}
\end{equation}
So, the learning algorithm has access to samples from the noisy distribution $\cd$, but needs to achieve good loss on the clean distribution $\cd^*$. We define a \emph{learner} in this setting to be an algorithm that minimizes (\ref{eq:loss}) using a finite number of samples:

\begin{definition}\label{def:learner}
For some learning algorithm $\ca$ and some distribution $\cd$ over $\cx \times \cy$, we say that $\ca$ is a \textbf{learner} for $\cd$ if there exists a function $m : (0,1) \to \naturals$ s.t. for every $\epsilon \in (0,1)$, taking $m \ge m(\epsilon)$ we get,
$\E_{S \sim \cd^{m}} L_{\cd^*}(\ca(S)) \le \epsilon$.

In this case, we call $m(\cdot)$ the sample complexity of $\ca$ w.r.t. $\cd$. Additionally, for some class $\cp$ of distributions over $\cx \times \cy$, we say that $\ca$ is a \textbf{learner} for $\cp$ if there is some $m(\cdot)$ s.t. $\ca$ is a learner with sample complexity $m(\cdot)$ for every $\cd \in \cp$. 
\end{definition}



Note that this definition of learner is similar to the notion of asymptotically consistent estimator. However, our definition explicitly accounts for the sample complexity. To give a concrete example, let us consider agnostic learning using the $\ERM$ rule, namely:
$\ERM_\ch(S) = \arg \min_{h \in \ch} L_S(h)$.

\label{def:p-h-gamma}For some hypothesis class $\ch$ and some margin $\gamma > 0$, let $\cp(\ch,\gamma)$ be the class of distributions such that for every $\cd \in \cp(\ch,\gamma)$ we have $\gamma(\cd) \ge \gamma$ and $f^*_\cd \in \ch$. Namely, $\cp(\ch,\gamma)$ is the class of distributions with margin $\gamma$ for which the Bayes optimal classifier comes from $\ch$.
The following theorem states that finite VC-dimension and a non-zero margin form a sufficient condition for learnability with noise.

\begin{theorem}
\label{thm:erm_learner}
There exists a constant $C > 0$ s.t. for every hypothesis class $\ch$ with $\VC(\ch) < \infty$, $\ERM_\ch$ is a \textbf{learner} for $\cp(\ch,\gamma)$ with sample complexity $m(\epsilon) =  C \frac{\VC(\ch) + \log(1/\epsilon)}{\epsilon^2\gamma^2}$.
\end{theorem}

The proof of Theorem~\ref{thm:erm_learner} is given in the appendix. The main idea is the following lemma (whose proof is also in the appendix), which shows that the margin assumption connects a small relative error w.r.t. $\cd$ to a small absolute error w.r.t. $\D^*$. 

\begin{lemma}
\label{lem:agnostic_loss_bound}
Fix some distribution $\cd$, and let $h^*$ be the Bayes optimal classifier for $\cd$. Assume that $\gamma_\delta(\cd) > 0$. 
Then, for every $h$ such that $L_\cd(h) \le L_\cd(h^*) + \epsilon$ it holds that
$L_{\cd^*}(h) \le \frac{\epsilon}{\gamma_\delta(\cd)} + \delta$.
\end{lemma}

Combining this lemma with the Fundamental Theorem of Learning Theory (see \cite{shalev2014understanding}) yields Theorem~\ref{thm:erm_learner}. From this theorem, we see that given more samples than the $\VC$-dimension (often corresponding to the number of parameters), learning is possible. However, complex classifiers with large $\VC$-dimension such as neural networks are often trained in the \textit{overparameterized} regime, when the number of parameters exceeds the number of samples. In this regime, the bound of Theorem \ref{thm:erm_learner} becomes vacuous. That said, this does not rule out the possibility of distribution-dependent sample complexity bounds using other measures of complexity (e.g., Rademacher complexity).

\subsection{Samplers}

As previously mentioned, 
neural networks trained on noisy data replicate the noise to unseen samples as well, behaving like \textit{samplers} from the noisy distribution. We next define formally a sampler for some distribution $\cd$, giving a similar definition as in \cite{nakkiran2020distributional}.

For some learning algorithm $\ca$, some number $m \in \naturals$ and some distribution $\cd$ over $\cx \times \cy$, define the distribution $\ca(\cd^m)$ over $\cx \times \cy$, where $(\x,y) \sim \ca(\cd^m)$ is given by sampling $S \sim \cd^m$, sampling $\x \sim \cd_\cx$ and setting $y = \ca(S)(\x)$.
Namely, $\ca(\cd^m)$ is the distribution given by (re)-labeling $\cd$ using a hypothesis generated by $\ca$ when observing a random sample of size $m$. Using these notations, we define a sampler algorithm for the distribution $\cd$ as follows:

\begin{definition}
For some learning algorithm $\ca$ and some distribution $\cd$ over $\cx \times \cy$, we say that $\ca$ is a \textbf{sampler} for $\cd$ if there exists $\widetilde{m} : (0,1) \to \naturals$ s.t. for every $\epsilon \in (0,1)$, taking $m \ge \widetilde{m}(\epsilon)$ we get,
$$\mathrm{TV}(\ca(\cd^m), \cd) \le \epsilon$$
where $\mathrm{TV}$ is the Total Variation Distance. Then, we call $\widetilde{m}$ the sample complexity of $\ca$ w.r.t. $\cd$. Additionally, for some class $\cp$ of distributions over $\cx \times \cy$, we say that $\ca$ is a \textbf{sampler} for $\cp$ if there exists $\widetilde{m}(\cdot)$ s.t. $\ca$ is a sampler with sample complexity $\widetilde{m}(\cdot)$ for each $\cd \in \cp$. \end{definition}

So, a sampler for $\cd$ is an algorithm that generates a distribution similar to $\cd$ (the noisy input distribution) when labeling new examples. A primary example for a sampler is the 1-Nearest-Neighbour algorithm: since 1-NN outputs the (possibly corrupted) label of the closest neighbour, its prediction behaves like sampling from  $\P_\cd(y|\x)$ (see Section \ref{sec:algorithms}). Figure~\ref{fig:intro} shows that neural networks behave similarly to samplers when trained on a noisy version of the CIFAR-10 dataset. 

Given the above definition, it can be easily shown that, instead of approximating the Bayes optimal prediction, a sampler preserves the noise rate of the original distribution.
\begin{lemma}
\label{lem:sampler-noise}
Let $\ca$ be a \textbf{sampler}  $\cd$ with sample complexity $\Tilde{m}$. Then, for $m\ge\Tilde{m}(\eps)$,
\[
\eta(\cd) - \eps \le \E_{S \sim \cd^m} L_{\cd^*}(\ca(S)) \le \eta(\cd) + \eps 
\]
\end{lemma}

While a \textit{sampler} for $\cd$ is not a good \textit{learner} (in the sense of Definition \ref{def:learner}), it does have some favorable properties. Primarily, it can take significantly fewer samples to get a sampler than it would take to get a learner, hence making the study of samplers more suitable for the overparameterized regime. In fact, in the extreme case one could get a sampler using only a \textit{single example} from the distribution, while getting a learner for the same distribution would require an arbitrarily large number of examples. Indeed, fix $b \in \{\pm 1\}$ and $\gamma \in (0,1)$, and let $\cd_b$ be the distributions concentrated on a single example $\x \in \cx$, with label $y \in \{\pm 1\}$ s.t. $\P_{\cd_b}(y = 1) = \frac{1+b\gamma}{2}$.
To get a sampler from $\cd_b$ it clearly suffices to take a single example $(\x,y)$, and return the constant function $y$. To find the Bayes optimal for the distribution $\cd_b$, on the other hand, any algorithm needs $\Omega(1/\gamma^2)$ examples. Using this observation, we show the folowing result:

\begin{theorem}
\label{thm:lower_bound}
For every $M > 0$, there exists a class of distributions $\cp_M$ such that:
\begin{itemize}
    \item There exists a \textbf{sampler} for $\cp_M$ with sample complexity $\widetilde{m}\equiv 1$.
    \item Any \textbf{learner} for $\cp_M$ has sample complexity satisfying $m(1/8) \ge M$.
\end{itemize}
\end{theorem}

\subsection{Teachers}
Motivated by the observed behavior of  neural networks in the overparemeterized regime, we defined the notion of \textit{samplers}. We showed that samplers can be much more sample efficient than learners, at the cost of giving noisy predictions. Next, we will show that while samplers are in and of themselves bad classifiers, they can still be good \textit{teachers}. That is, we can use samplers to label a large unlabeled dataset and train a \textit{student} classifier on this new dataset. This process is often referred to as \textit{knowledge distillation}, and it has been shown to work remarkably well in practice \citep{pseudolabels2019,metapsedulabels2020}. The following definition captures the notion of a (good) teacher:

\begin{definition}
\label{def:teacher}
For some learning algorithm $\ca$, some distribution $\cd$ over $\cx \times \cy$, we say that $\ca$ is a \textbf{teacher} for $\cd$ if there exists a function $\widetilde{m}:(0,1)^2 \to \naturals$ s.t. for every $\epsilon,\tau \in (0,1)$, taking $m \ge \widetilde{m}(\epsilon,\tau)$ the following holds:
\begin{enumerate}
    \item $L_{\cd^*}\left(f^*_{\ca(\cd^m)}\right) = \prob{\x \sim \cd}{f^*_\cd(\x) \ne f^*_{\ca(\cd^m)}(\x)} \le \epsilon$
    \item $\gamma_\epsilon(\ca(\cd^m)) \ge \gamma(\cd) - \tau$
\end{enumerate}

Additionally, for some class $\cp$ of distributions over $\cx \times \cy$, we say that $\ca$ is a \textbf{teacher} for $\cp$ if there is some $\widetilde{m}(\cdot, \cdot)$ s.t. $\ca$ is a teacher with sample complexity $m(\cdot, \cdot)$ for every $\cd \in \cp$.
\end{definition}

The first condition in the above definition means that the Bayes optimal of the original distribution and the distribution induced by the teacher are close. The second condition means that the probability mass of low margin samples from the distribution labeled by the teacher is small. Intuitively, an algorithm satisfying Definition \ref{def:teacher} is a good teacher since the distribution it induces is ``similar'' to the original distribution. Hence, if the student finds a good hypothesis on the teacher-induced distribution, its hypothesis is also good w.r.t. the original distribution. 

In Section \ref{sec:distillation} we formally study when and how teachers can be used for distillation, giving guarantees for getting students with small error with respect to the clean distribution $\cd^*$. Before that, let us first show that samplers are indeed good teachers.

\begin{theorem}
\label{thm:sampler_teacher}
Let $\ca$ be a \textbf{sampler} for $\cd$ with sample complexity $m(\cdot)$ and margin $\gamma$. Then, $\ca$ is a \textbf{teacher} for $\cd$ with sample complexity $\widetilde{m}(\epsilon, \tau) = m(\eps\cdot\min(\tau/2, \gamma))$.
\end{theorem}
The main idea behind the proof of Theorem~\ref{thm:sampler_teacher} is that, given an example with probability mass $p$, the ``cost" (in terms of $\mathrm{TV}$) of flipping the example's label w.r.t. the Bayes optimal $f^*_\cd$ is at least $p\cdot \gamma$. Since our TV ``budget" is limited by $\eps\gamma$ we can only flip a probability mass of $\eps$ of the distribution.


Finally, before moving on to discuss the implication of our results, we show that learners are also good teachers. This is almost immediate, since learners approximate the Bayes optimal classifier, and hence can be used to train students to similarly imitate the Bayes classifier.

\begin{theorem}
\label{thm:learner_teacher}
Let $\ca$ be a \textbf{learner} for $\cd$ with sample complexity $m(\cdot)$. Then, $\ca$ is a \textbf{teacher} for $\cd$ with sample complexity $\widetilde{m}(\epsilon,\tau) = m\left(\frac{\epsilon(1-\gamma(\cd)+\tau)}{2}\right)$.
\end{theorem}

To conclude, Theorem \ref{thm:sampler_teacher} and Theorem \ref{thm:learner_teacher} show that, to some extent, a teacher is an ``interpolation'' between a sampler and a learner.


\section{Distillation from Teachers}
\label{sec:distillation}
We defined the notion of a teacher, and showed that both samplers and learners are teachers. Now, we show how teachers (and in particular, samplers) can be used to find good learners. First, we show that an ensemble of teachers can be used to get a good learner by simply outputting the majority vote of the ensemble. Next, we show that using the ensemble to label a new set of unlabeled examples (i.e., performing knowledge distillation) guarantees finding a student with small loss, assuming the Bayes optimal classifier comes from the hypothesis class learned by the student. Such process is favorable, since it reduces the computational cost of running the ensemble at inference time, and also allows using a different hypothesis class for the student (for example when using a student network of smaller size,  e.g., \cite{gou2021knowledge}).
Finally, we show that distillation can also be achieved by labeling examples using a teacher that is randomly chosen from a fixed set of teachers, a method that has some computational benefits at training time. To the best of our knowledge, this is a novel technique for distillation that has not been previously suggested in prior work.

\subsection{Ensembles of Teachers}

We now show how an ensemble of teachers can be used to get accurate predictions with respect to the Bayes optimal predictor.
Given some $k$ samples $S_1, \dots, S_k \sim \cd^m$, each one of size $m$, we can use a learning algorithm $\ca$ to get $k$ different hypothesis $h_1, \dots, h_k$, where $h_i := \ca(S_i)$. Observe the ensemble hypothesis, which outputs the majority vote of the ensemble members:
\[
h_{\ens}(\x) = \arg \max_{y \in \cy} \sum_{i} \ind\{h_i(\x) = y\}
\]
We use the notation $\ca_{\ens}(S_1, \dots, S_k) := h_{\ens}$ to denote this ensemble hypotheses. The following Theorem states that the ensemble hypothesis has a good loss on average, when using a large enough ensemble of teachers:

\begin{theorem}
\label{thm:ensemble_inference}
Assume that $\ca$ is a \textbf{teacher} for some distribution $\cd$ with complexity $\widetilde{m}$. Then, for all $\epsilon \in (0,1)$, taking $m \ge \widetilde{m}\inparen{\frac{\epsilon}{3},\frac{\gamma(\cd)}{2}}$ and $k \ge 
\frac{16 \log(3/\epsilon)}{\gamma(\cd)^2}$ we get,
$$\E_{S_1, \dots, S_k \sim \cd^m} L_{\cd^*}(\ca_\ens(S_1, \dots, S_k)) \le \epsilon$$
\end{theorem}

We now give a sketch of the proof. For some $\x \in \cx$, let $y_i$ be the prediction of the $i$-th teacher, and let $\bar{y}$ be the average of the predictions, namely $\bar{y} = \frac{1}{k} \sum_i y_i$. Observe that $h_\ens(\x) = \sign(\bar{y})$, and additionally $\E [\bar{y}] = \E_{\ca(\cd^m)}[y|\x]$. Now, since $\ca$ is a teacher we get $\E_{\ca(\cd^m)} [y|\x] \approx \E_\cd [y|\x]$ (with high probability over the choice of $\x$), and by concentration bounds this implies that w.h.p. $h_\ens(\x)$ give the Bayes optimal prediction (i.e., $h_\ens(\x) = f^*_\cd(\x)$).

So, Theorem \ref{thm:ensemble_inference} shows that if $\ca$ is a \textbf{teacher} for $\cd$, then $\ca_\ens$ with $k \ge \widetilde{\Omega}\inparen{1/\gamma(\cd)^2}$\footnote{We use $\tilde{\Omega}$ to hide constant and logarithmic factors.} is a \textbf{learner} for $\cd$.
More generally, if we have a teacher for some distribution $\cd$ with positive margin, this implies that there exists a learner for the same distribution. The inverse of this statement gives another interesting result---if no algorithm can learn some problem, then getting a teacher (or sampler) is also hard. Formally, let $\cp$ be a class of distributions over $\cx \times \cy$ s.t. for all $\cd \in \cp$ it holds that $\gamma(\cd) \ge \gamma$. Then, if there is no \textbf{learner} for $\cp$, there is no \textbf{teacher} or \textbf{sampler} for $\cp$.
Additionally, a similar result holds for problems which are computationally hard to learn. That is, if there is no learner for $\cp$ that runs in polynomial time, then there is no poly-time teacher or sampler for $\cp$. Observe that the condition that $\gamma(\cd) \ge \gamma$ for all $\cd \in \cp$ in the previous statement is necessary. Indeed, taking $\cp = \cup_{M=1}^\infty$, where $\cp_M$ is the distribution class guaranteed by Theorem \ref{thm:lower_bound}, gives a class $\cp$ s.t. there is a sampler for $\cp$ with sample complexity $\widetilde{m} \equiv 1$, but there is no learner for $\cp$.


\subsection{Distillation from Ensembles}

We showed that an ensemble of $k=\Tilde{\Omega}(\gamma^{-2})$ teachers gives a classifier that approximates the Bayes optimal predictor. This, however, incurs a $k$ factor in computational cost at inference time. To prevent this, we can instead use the ensemble to label new unlabeled data, and train a new classifier to imitate the ensemble. This moves the computational burden from inference time to training time.

Fix some class $\ch$, and define the \textbf{Ensemble-Pseudo-Labeling (EPL)} algorithm as follows:
\begin{enumerate}[topsep=4pt,partopsep=0pt,itemsep=4pt,parsep=0pt,leftmargin=32pt]
    \item For some $k, m \in \naturals$, sample $S_1, \dots, S_k \sim \cd^m$.
    \item Run $\ca$ on $S_1, \dots, S_k$, and let $h_\ens = \ca_\ens(S_1, \dots, S_k)$.
    \item Take $S'$ to be a set of $m'$ \textbf{unlabeled} examples sampled from $\cd_\cx$, and label it using $h_\ens$. 
    \item Denote by $\widetilde{S}$ the pseudo-labeled set. Run $\ERM_\ch$ on the set $\widetilde{S}$ and return $h := \ERM_\ch(\widetilde{S})$.
\end{enumerate}

Intuitively, since $h_\ens$ approximates the Bayes optimal classifier (see Theorem \ref{thm:ensemble_inference}), the labels for the new dataset $S'$ are mostly correct. In other words, the pseudo-labeled set $\widetilde{S}$ comes from a distribution that is close to the clean distribution $\cd^*$. When using pseudo-labels, it is enough to use unlabeled data, which is often abundant, so $\widetilde{S}$ can be much larger than our original labeled dataset. In this case, we no longer need to work in the overparameterized regime, so $\ERM_\ch$ is guaranteed to achieve good performance by standard $\VC$ bounds. This argument is captured in Theorem~\ref{thm:ensemble_pseudo_labeling}:
\begin{theorem}
\label{thm:ensemble_pseudo_labeling}
For hypothesis class $\ch$ with $\VC(\ch) < \infty$, and let $\cd \in \hyperref[def:p-h-gamma]{\cp(\ch,\gamma)}$ for some $\gamma > 0$.
Let $\ca$ be a \textbf{teacher} for $\cd$ with distributional sample complexity $\widetilde{m}$. Then, there exists a constant $C > 0$ s.t. for every $\epsilon \in (0,1)$, running the \textbf{EPL} algorithm with parameters $m \ge \widetilde{m}\inparen{\frac{\epsilon}{12}, \frac{\gamma}{2}}$, $m' \ge C \frac{\VC(\ch) + \log(1/\epsilon)}{\epsilon^2}$ and $k \ge \frac{16 \log(12/\epsilon)}{\gamma^2}$ returns a hypothesis $h$ satisfying
$\E_{S_1, \dots, S_k, \widetilde{S}} L_{\cd^*}(h) \le \epsilon$.
\end{theorem}




\subsection{Distillation from Random Teachers}

While the \textbf{EPL} algorithm moves the computation cost from inference to training, we still suffer a $k$ factor for labeling each example. Instead, a possible solution is to label examples by choosing a random classifier from the ensemble. Then, we only run one classifier per example. So, we define the \textbf{Random-Pseudo-Labeling (RPL)}  similarly to \textbf{EPL}, except that for each example $\x$ in the unlabeled dataset $S'$, we pick $h \sim \{h_1, \dots, h_k\}$ and label $\x$ by $h(\x)$.  

To understand why this method works, we can think of $\widetilde{S}$ as coming from a distribution $\widetilde{\cd}$, defined by sampling $\x \sim \cd_\cx$ and sampling $y$ s.t. $\E_{\widetilde{\cd}}[y|\x] = \E_{i \sim [k]}[h_i(\x)]$. By the properties of the teacher, using concentration arguments as in Theorem \ref{thm:ensemble_inference}, the distribution $\widetilde{\cd}$ is close to the \textit{noisy} distribution $\cd$. However, as mentioned before, the advantage of using $\widetilde{S} \sim \widetilde{\cd}$ is that we can use a much larger set of unlabeled data, in which case the result of Theorem \ref{thm:erm_learner} can be applied to show that the above algorithm finds a hypothesis with good error. This is stated in the following result:

\begin{theorem}
\label{thm:random_ensemble}
For hypothesis class $\ch$ with $\VC(\ch) < \infty$. let $\cd \in \hyperref[def:p-h-gamma]{\cp(\ch,\gamma)}$ for $\gamma > 0$.
Let $\ca$ be a \textbf{teacher} for $\cd$ with distributional sample complexity $\widetilde{m}$. Then, there exists a constant $C > 0$ s.t. for every $\epsilon \in (0,1)$, running the \textbf{RPL} algorithm with parameters $m \ge \widetilde{m}\inparen{\frac{\epsilon\gamma}{54}, \frac{\gamma}{2}}$, $m' \ge C \frac{\VC(\ch) + \log\inparen{\epsilon^{-1} \gamma^{-1}}}{\epsilon^2\gamma^2}$ and $k \ge \frac{128 \log\inparen{36\epsilon^{-1} \gamma^{-1}}}{\gamma^2}$ returns $h$ satisfying
$\E_{S_1, \dots, S_k, \widetilde{S}} L_{\cd^*}(h) \le \epsilon$.
\end{theorem}

Compare the sample complexity of the above theorem with the sample complexity achieved by the EPL algorithm, stated in Theorem \ref{thm:ensemble_pseudo_labeling}. At first glance, it seems that the gain from using the RPL algorithm is not clear, as it increases the number of unlabeled data by a factor of $1/\gamma(\cd)^2$ (and also might increase the number of labeled examples). This is not surprising, because a dataset labeled by a random classifier can be much more noisy than a dataset labeled by the ensemble, and hence more examples are required in order to learn. Note, that since $k \ge \tilde{\Omega}(1/\gamma(\cd)^2)$, the randomized labeling takes $1/k$ compute per example relative to the ensemble labeling, it will, however, need to label an order of $k$ times more examples.

We do note that there might still be computational benefits to using the random approach, as it allows the training and labeling to happen in parallel without a significant increase in compute (e.g., 2 GPU cores are enough to train in parallel without any loss of time). On the other hand, labeling a dataset using the ensemble requires one to either increase the compute by $k$ (applying parallelism), or otherwise wait until the full dataset is labeled, and only then start the distillation process. Finally, our experiments show that the gain from ensemble labeling over random labeling, as captured by the final accuracy of the trained student, is not so significant (see Table~\ref{tab1}). This suggests that in some particular cases (e.g., under further assumptions on the data distribution and/or the optimization algorithm), it is preferable to use the RPL algorithm.

\section{Samplers Exist}
\label{sec:algorithms}

So far, we showed that samplers can be good teachers. These can then be used for knowledge distillation, generating students which approximate the Bayes optimal prediction.
To complete the picture, we now show that some well-known algorithms can be used to get samplers or teachers. We start by studying the k-Nearest-Neighbour algorithm, and show that it is a teacher, when the underlying distribution has some Lipschitz property.
We then analyze Lipschitz classes of functions (e.g., linear functions, kernels and neural networks), and show that when the data is well-clustered, these algorithms can be teachers as well. In all cases, the sample complexity needed for the samplers or teachers is lower than the complexity required for learning.

\subsection{Nearest Neighbour Algorithm}

For some set $S = \{(\x_1, y_1), \dots, (\x_m, y_m)\} \subseteq \cx \times \cy$ we denote $S_\cx = \{\x_1, \dots, \x_m\}$. Fix some metric $d$ over the space $\cx$. In this section, we assume that the metric space $(\cx,d)$ satisfies the Heine-Borel property---that is, every closed and bounded set in $\cx$ is compact\footnote{Specifically, the Heine-Borel property holds for $\reals^n$ where $d$ is induced by some norm.}.
For some finite set $S \subseteq \cx \times \cy$, and some $\x \in \cx$, denote $d(\x,S) := \min_{\x' \in S_\cx} d(\x,\x')$ and $\pi(\x,S) := \arg \min_{(\x',y') \in S} d(\x,\x')$. Additionally, we define the set $\kpi(\x, S_\cx) \subseteq S$ to be the set of $k$ points in $S$ that are closest to $\x$. That is, $\kpi(\x,S)$ is a set of size $k$ s.t. for any $(\x',y') \in \kpi(\x,S)$ and $(\tilde{\x}, \tilde{y}) \in S \setminus \kpi(\x,S)$ it holds that $d(\x,\x') \le d(\x,\tilde{\x})$\footnote{If there are multiple choices for such set, we choose one arbitrarily.}.
For some distribution $\cd$, we say that $\cd$ is $\lambda$-Lipschitz if for all $\x, \x' \in \supp(\cd_\cx)$ and $y \in \cy$ it holds that
$\abs{\P_\cd[y|\x] -\P_\cd[y|\x']} \le \lambda d(\x,\x')$.

For some odd $k \ge 1$, define
$\ca_{\kNN}(S)(\x) := \arg \max_{\hat{y} \in \cy}
\abs{(\x',y') \in \kpi(\x,S),~y'=\hat{y}}$, i.e., $\ca_{\kNN}(S)$ is the k-NN algorithm over sample $S$.
We start by showing that the $\ca_{\oNN}$ is a sampler.
The distributional sample complexity of $\ca_{\oNN}$ depends on the number of $\epsilon$-balls that can cover a $1-\delta$ mass of the distribution (Lemma \ref{lem:covering} in the Appendix shows that this number is always finite).
Given such cover, we can guarantee that with a large enough sample, we find a candidate example in each of the balls that have non-negligible mass. In that case, if the distribution of labels does not change significantly in each ball, $\ca_{\oNN}$ is indeed a sampler:

\begin{theorem}
\label{thm:one_nearest_neighbour}
Let $\cd$ be some $\lambda$-Lipschitz distribution. Then, $\ca_{\oNN}$ is a \textbf{sampler} for $\cd$.
\end{theorem}

Next, we study the k-Nearest-Neighbour algorithm for any odd $k \ge 1$. Similarly to the analysis of the 1-Nearest-Neighbour case, we show that a large enough sample guarantees at least $k$ candidates in each of the $\epsilon$-ball covering the distribution. In this case, the prediction of the k-Nearest-Neighbour algorithm is the majority vote over the $k$ neighbours in each ball. This, in fact, will grow closer to the Bayes optimal prediction as $k$ grows, and therefore k-Nearest-Neighbour algorithm is not strictly speaking a sampler for $k > 1$. However, we show that it is always a teacher:

\begin{theorem}
\label{thm:knn}
Let $\cd$ be some $\lambda$-Lipschitz distribution. Then, $\ca_{\kNN}$ is a \textbf{teacher} for $\cd$.
\end{theorem}

The core argument for proving Theorem \ref{thm:knn}, beyond the covering argument used in the 1-Nearest-Neighbour case, relies on using a variant of Condorcet's Jury (CJT) Theorem. Roughly speaking, the theorem states that the accuracy of the majority vote of a set of predictors is better than the average accuracy of the individual predictors. In the k-Nearest-Neighbour case, each of the $k$ candidates casts a ``vote'', and using CJT we show that this improves over the 1-Nearest-Neighbour prediction, which is already a sampler (and hence a teacher). Notice that this works for any $k\ge 1$, and we do not need to require $k$ to be large enough, as would be required in order to get a learner.

\paragraph{Case Study: Limited Memory 1-NN.}
Now, we introduce a case study where applying the distillation methods studied in Section \ref{sec:distillation} results in a low-error student classifier, while at the same time using the same learning algorithm on the labeled data alone would maintain high levels of noise in the prediction. While this example is somewhat synthetic, we believe that it captures the behavior of samplers and teachers in practice. In this part, we take $\cx = \{0,1\}^n$ to simplify the analysis\footnote{Note that similar arguments can be applied to the case where $\cx = \reals^n$.}.

Given our previous results, taking the 1-Nearest-Neighbour algorithm for the teacher seems to be a reasonable choice. However, using 1-NN as a student is problematic, since the $\VC$-dimension of 1-NN classifiers is infinite. Thus, we consider a similar hypothesis class of 1-NN with limited memory.
Let $\ch_b$ be the class of 1-NN predictors with memory of size $b$. That is, for every $h \in \ch_b$ there is some $S \subseteq \cx \times \cy$ such that $S$ can be stored using $b$ bits of memory, and for every $\x \in \cx$ we have $h(\x) = \hat{y}$ where $\hat{y}$ is the label of the closest neighbour to $\x$ in $S$, namely $(\hat{\x}, \hat{y}) = \pi(\x,S)$. Indeed, observe that $\ch_b$ is a finite class of size $2^b$, and therefore $\VC(\ch_b) \le \log(\abs{\ch_b}) = b$. 

So, consider the following setting: let $\cd$ be a distribution s.t. $f^*_\cd \in \ch_b$. Assume that we draw a labeled dataset of size $k m$ from $\cd$ ($k$ subsets of size $m$), and that $km \cdot n< b$ (that is, we assume that the teacher is trained in the overparameterized regime). In this case, we can use $\ca_{\oNN}$ as the algorithm for the teacher, and $\ERM_{\ch_b}$ as the student.\footnote{Since we assume the teacher is trained in the overparameterized regime, there can be many $\ERM$ solutions, and we need to specify which one is chosen, so simply taking $\ERM_{\ch_b}$ is not enough. A natural choice in the overparameterized regime is to simply use $\ca_{\oNN}$.} Indeed, from everything we showed so far it follows that both the EPL and the RPL algorithms will yield a student which approximates the Bayes optimal predictor. Additionally, observe that by Theorem \ref{thm:one_nearest_neighbour}, simply using $\ca_{\oNN}$ over the entire training set will yield a sampler, which can be far from the Bayes optimal predictor (see Lemma \ref{lem:sampler-noise}).

Admittedly, one could always use k-NN (instead of 1-NN) and get a classifier that approximates the Bayes optimal prediction, when $k$ is sufficiently large. 
However, in practice, ``black-box'' algorithms such as neural networks can \textit{behave like} 1-NN (as shown in \cite{nakkiran2020distributional}). In this case, what we show is essentially that knowledge distillation can take an algorithm that \textit{behaves like} 1-NN and turn it into an algorithm that \textit{behaves like} k-NN, without knowledge of the internals of the algorithm, and without suffering additional computational costs at inference time.

\subsection{Bounded-Norm Infinite-Width ReLU Networks}

We continue with investigating infinite-width neural-network with weights of bounded norm, following the setting studied in \cite{savarese2019infinite, ongie2019function}. We define a ReLU network of width $k$ and depth 2 by:
\[
h_{\theta}(\x) = \sum_{i=1}^k w_i^{(2)} \sigma \inparen{\inner{\bw_i^{(1)}, \x} + b_i^{(1)}} + b^{(2)}
\]

where $\theta = (k,W^{(1)}, W^{(2)}, b^{(1)}, b^{(2)})$, and $\sigma$ is the ReLU activation, namely $\sigma(x) = \max \{x,0\}$.
As in \cite{savarese2019infinite}, we consider the Euclidean norm of non-biased weights:
\[
C(\theta) = \frac{1}{2} \sum_{i=1}^k \inparen{\inparen{w_i^{(2)}}^2 + \norm{\bw_i^{(1)}}_2^2}
\]
Now, consider fitting a sample $S \subseteq \cx \times \cy$ with a network $h_\theta$ with $C(\theta)$ acting as regularization. Namely, observe the following objective function:
\[
R(S) = \inf_\theta C(\theta) ~\mathrm{s.t.}~ h_\theta(\x) = y~\textforall~(\x,y) \in S
\]

In the one-dimensional case, i.e. when $\cx = \reals$, Theorem 3.3 from \cite{savarese2019infinite} shows that $R(S)$ gives the linear spline interpolation of the data points. Namely, let $\hat{\theta} := R(S)$, and assume that $S = \{(x_1, y_1), \dots, (x_m, y_m)\}$ is sorted such that $x_1 < x_2 < \dots < x_m$ (assuming there are no repeated samples). Then, for every $i \in [m]$ and for all $x \in [x_i,x_{i+1}]$ it holds that
\[
h_{\hat{\theta}}(x) = y_i + \frac{y_{i+1}-y_i}{x_{i+1}-x_i}\inparen{x-x_i}
\]

In this case, it can be easily shown that for all $x \in [x_1, x_m]$ we have $\sign h_{\hat{\theta}}(x) = \oNN(S)(x)$, so training a network with bounded-norm weights (and unbounded width) behaves like nearest neighbour classification over the range covered by the sample. Using this, we show that ReLU networks in this setting are samplers:
\begin{theorem}
\label{thm:bounded_norm_relu}
Let $\ca$ be the algorithm that takes a sample $S$ and returns a function $h$ s.t. $h(x) =  \sign \inparen{h_{\hat{\theta}}(x)}$, for $\hat{\theta} = R(S)$. Let $\cd$ be some continuous\footnote{We define a continuous distribution over $\reals$ to be any distribution s.t. the function $f(a) = \P[x \ge a]$ is continuous. Observe that in this case, w.p. $1$ there are no repeated examples in $S$, therefore avoiding the case where $R(S)$ is not well-defined.} $\lambda$-Lipschitz distribution. Then, $\ca$ is a \textbf{sampler} for $\cd$.
\end{theorem}

The above shows that when we don't limit the size of the network (i.e., in the overparameterized regime), and use the weights' norm as regularization, the resulting algorithm is a sampler. Observe that using such regularization is necessary for this result. That is, simply choosing \textit{some} network that fits the data (rather than choosing the network with minimal norm), does not necessarily give a sampler. Indeed, observe that one can construct a ReLU network $h_\theta$ that outputs a constant value (e.g., $1$) for all $x \in \reals$, outside of infinitesimally small neighbourhoods of the points of $S$, where $h_\theta$ interpolates the data (namely, $h_\theta$ is constant with very narrow ``spikes'' towards the correct labels of the examples in the sample). Thus, on new points $h_\theta$ evaluates to $1$ with high probability, so it does not behave like a sampler.

Admittedly, going beyond the one-dimensional case is more challenging, as it requires understanding the high-dimensional geometry of the function returned by $R(S)$. While we defer this case for future work, we note that the analysis of \cite{ongie2019function} gives some technical tools for understanding the multivariate version of the above problem. Specifically, \cite{ongie2019function} show that solving $R(S)$ is equivalent to minimizing a specific norm in function-space, which controls the complexity of the learned function. Among other things, the authors show that controlling this norm prevents a ``spiking'' behavior as described above.

\subsection{Well-Clustered Data and Lipschitz Classes} 
We now show that under certain clustering assumptions, many learning methods can be teachers. First, we study a simplified case of a distribution supported on a finite set. The following theorem shows that when the hypothesis class shatters the support of the distribution, $\ERM_\ch$ is a teacher with sample complexity $\tilde{O}(k/\epsilon)$.

\begin{theorem}
\label{thm:finite-classes}
Fix some hypothesis class $\ch$, and let $\cd$ be some distribution over $\cx \times \cy$ such that $\abs{\supp(\cd_\cx)} = k \le \VC(\ch)$ and the support of $\cd_\cx$ is shattered. Then, $\ERM_\ch$ is a teacher, with sample complexity $\widetilde{m}(\epsilon,\tau) = \frac{2k\log(2k/\eps)}{\eps}$.
\end{theorem}
Contrast this with Theorem~\ref{thm:erm_learner}, where we show that when $\VC(\ch)=d$, $\ERM$ is a learner with sample complexity $m(\eps)=\tilde{O}\inparen{\frac{\VC(\ch)}{\epsilon^2\gamma(\cd)^2}}$.
This shows that sampling can be achieved in this case without a dependence on $1/\gamma^2$, as would be needed in order to get a learner. In fact, Theorem \ref{thm:lower_bound} shows that the dependence on $1/\gamma^2$ in the sample complexity of a learner cannot be avoided.

We proceed to discuss a more general version of Theorem~\ref{thm:finite-classes} where $\cd$ is well-clustered in $k$ balls of small radius (similar to a Mixture of Gaussians with low variance). In this case, we study $L$-Lipschitz hypothesis classes, defined as follows:

\begin{definition}
A hypothesis class $\ch$ is $L$-Lipschitz if for every $h \in \ch$ there exists some $\hat{h} : \cx \to \reals$ such that  $\hat{h}$ is $L$-Lipschitz and $h(\x) = \sign \hat{h}(\x)$ for all $\x \in \cx$.
\end{definition}

We note that a large family of learning methods such as bounded norm linear classifiers, kernel machines and shallow neural networks with Lipschitz activations (e.g., ReLU) are Lipschitz classes.
For learning $L$-Lipschitz classes, we study the $\ERM$ rule with respect to the hinge-loss (over the real-valued output) instead of the zero-one loss. Namely, we define $\ERM_\ch^{hinge}(S) = \arg \min_{h \in \ch} \E_{(\x,y)} \left[\ell_{\mathrm{hinge}}\inparen{y, \hat{h}(\x)}\right],$ where $\ell_{\mathrm{hinge}}(y,\hat{y}) = \max(1-y\hat{y}, 0)$.

We use the hinge-loss as it is often required that the output of a real-valued hypothesis separates the data with some margin. Indeed, since the zero-one loss is invariant to scale, the $L$-Lipschitz assumption under the zero-one loss is meaningless, since the hypothesis can always be scaled down to satisfy any Lipschitz bound. So, when the data is well-clustered and the hypothesis class $\ch$ is $L$-Lipschitz, $\ERM_\ch^{hinge}$ is a teacher with sample complexity $\Tilde{O}(\frac{k}{\gamma^2\eps})$. While this bound does depend on $1/\gamma^2$, it still improves the sample complexity of learning derived from Theorem \ref{thm:erm_learner}.
  
\begin{theorem}
\label{thm:lipschitz_class}
For $L$-Lipschitz class $\ch$, and some $\lambda$-Lipschitz distribution $\cd$  s.t. $\supp(\cd_\cx) \subseteq \cup_{i=1}^k B(\bc_i, r)$, where $r=\frac{\gamma}{2\max(\lambda, 3L)}$ and $k \le \VC(\ch)$ so the set of balls $B(\bc_i, r)$ can be shattered. Then, $\ERM_\ch^{hinge}$ is a teacher, with sample complexity $\widetilde{m}(\epsilon,\tau) = \tilde{O}(\frac{k\log(2k/\eps)}{
\gamma^2\eps })$.

\end{theorem}

\begin{figure}[t]\hspace{-0.4cm}
\begin{floatrow}
\ffigbox{%
  \centering
    \includegraphics[width=0.46\textwidth]{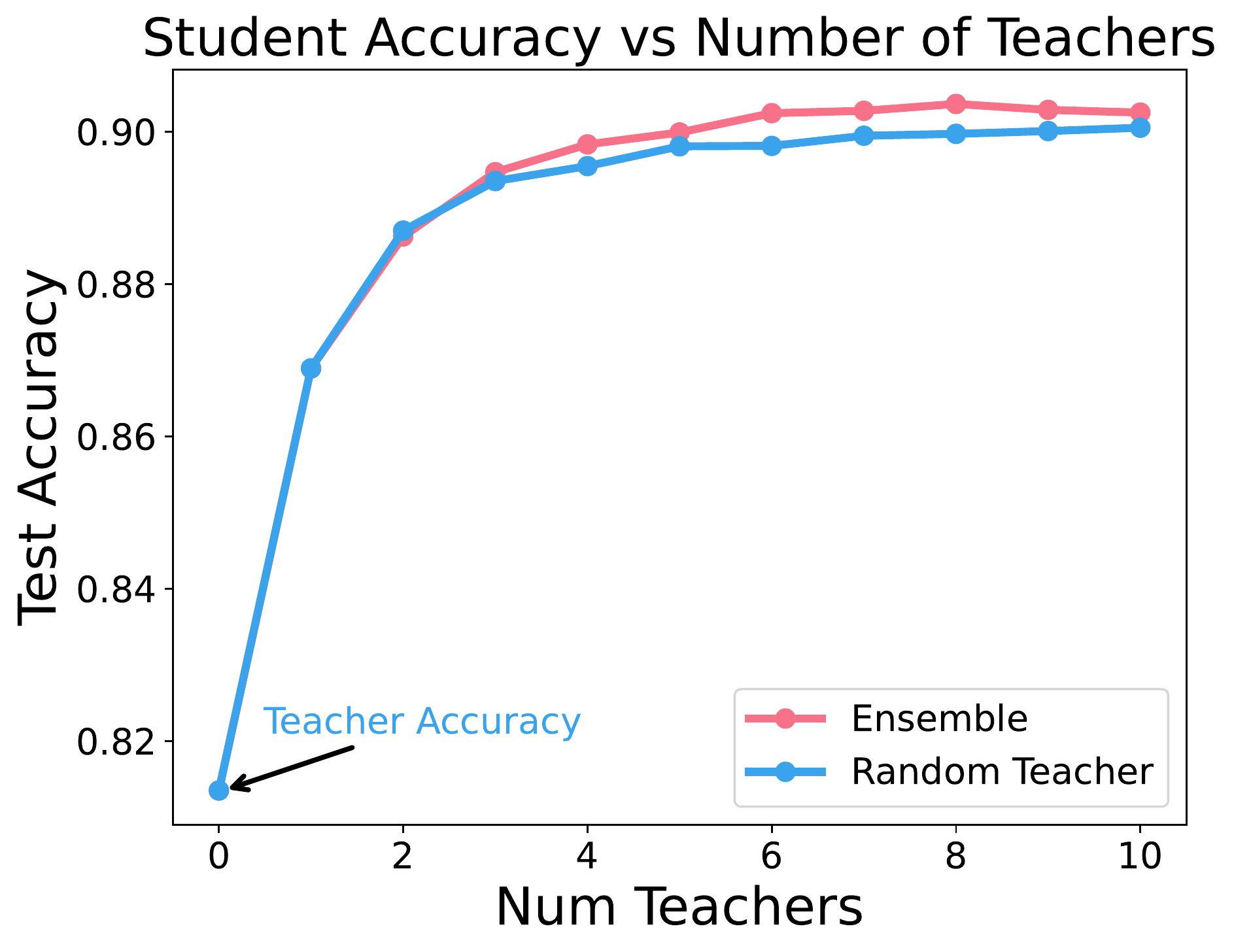}\vspace{-0.3cm}
}{%
  \caption{The effect of the number of teachers on the performance of the student.
  }\label{fig:num-teachers}%
}\hfill
\capbtabbox{%
\centering
\begin{tabular}[ht]{lc} \toprule
Experiment &  Test Accuracy  $\pm$ std \\
\midrule
\midrule
One Teacher & 0.868 $\pm 5$e-$3$  \\
5 Random Teachers & 0.898 $\pm 2$e-$3$ \\
10 Random Teachers & 0.900  $\pm 2$e-$3$ \\
5 Teacher Ensemble & 0.899  $\pm 2$e-$3$ \\
10 Teacher Ensemble & 0.902 $\pm 1$e-$3$ \\
\midrule
10-Ensemble Inference & 0.878 \\
10-Teacher Clean Ens. & 0.934  $\pm 0.8$e-$3$ \\
Teacher Accuracy & 0.813 $\pm 4.7$e-$2$\\
\bottomrule
\end{tabular}
}{%
  \caption{Comparison of teachers, students and ensembles test performance.
  }\label{tab1}
}
\end{floatrow}
\vspace{-0.7cm}
\end{figure}

\section{Experiments}
\label{sec:experiments}
To this point, we saw that getting a sampler (and thus a teacher) from a noisy distribution can be more sample efficient than getting a learner. Furthermore, we showed that we can leverage multiple independent teachers to approximate the Bayes optimal classifier either via ensembling at inference time or via distillation on unlabeled data. We now complement our theoretical results with an experimental evaluation, showing the benefit of using distillation when training on noisy data.
While in our theoretical setting we studied teachers that are trained on entirely disjoint training sets, in practice we find it more effective to train the teachers on overlapping datasets, as well as training on same dataset with different random initialization.



To get the teachers, we train a ResNet-18 \citep{he2015} on CIFAR-10 with 20\%-\emph{fixed} and \emph{non-uniform} label noise (see full details in \ref{sec:exp-app}). We see that our teachers achieve $81.3\%$ test accuracy (see Table~\ref{tab1}) and behave closely to samplers (see Figure~\ref{fig:intro}) reproducing the results of \citet{nakkiran2020distributional}. We now compare the three methods considered before for using teachers to get learners: 1) Test time Ensembling; 2) Ensemble as distillation teacher and; 3) Random teacher distillation. For distillation, we train a student network on the CIFAR-5m, a large (5-million examples) dataset that resembles the CIFAR-10 dataset \citep{cifar5m}, where the labels are provided by the previously trained teachers. We report our results in Table~\ref{tab1}, where the reported accuracies are on the CIFAR-10 test data. Observe that using an ensemble for inference reduces the noise significantly, and achieves test accuracy of $87.8\%$ (versus $81.3\%$ for a single teacher). When applying distillation, both random pseudo-labeling and ensemble pseudo-labeling further increase the test accuracy to about $90\%$. In addition, we study how the number of teachers affects performance (see Figure~\ref{fig:num-teachers}). We observe that both random pseudo-labeling and ensemble majority improve in performance when the number of teachers grow.


\newpage
\bibliography{refs}
\newpage
\appendix

\section{Proofs for Section~\ref{sec:setting}}

\begin{proof}{of Lemma \ref{lem:agnostic_loss_bound}.}

For every $\x$ denote $\gamma(\x) = \prob{}{y = h^*(\x)|\x}-\prob{}{y \ne h^*(\x)|\x}$, and since $h^*$ is the Bayes optimal predictor it holds that $\gamma(\x) \ge 0$.
Observe the following:
\begin{equation}
\label{eq:alb_1}
\begin{split}
    L_{\cd}(h) &= \mean{\x \sim \cd}{\prob{}{y \ne h(\x)|\x}} \\
    &= \mean{\x\sim\cd}{\prob{}{y \ne h(\x)|\x} \cdot \ind\{h(\x) = h^*(\x)\}}\\&+ \mean{\x\sim\cd}{\prob{}{y \ne h(\x)|\x} \cdot \ind\{h(\x) \ne h^*(\x)\}}\\
    &= \mean{\x\sim\cd}{\prob{}{y \ne h^*(\x)|\x} \cdot \ind\{h(\x) = h^*(\x)\}} \\&+ \mean{\x\sim\cd}{\prob{}{y = h^*(\x)|\x} \cdot \ind\{h(\x) \ne h^*(\x)\}} \\
    &= \mean{\x\sim\cd}{\prob{}{y \ne h^*(\x)|\x} \cdot \inparen{\ind\{h(\x) = h^*(\x)\} + \ind\{h(\x) \ne h^*(\x)\}}} \\&+ \mean{\x\sim\cd}{\gamma(\x) \cdot \ind\{h(\x) \ne h^*(\x)\}} \\
    &= L_\cd(h^*) + \mean{\x\sim\cd}{\gamma(\x) \cdot \ind\{h(\x) \ne h^*(\x)\}}
\end{split}
\end{equation}
Now, notice that we have:
\begin{equation}
\label{eq:alb_2}
\begin{split}
    \mean{\x \sim \cd}{\gamma(\x) \cdot \ind\{h(\x) \ne h^*(\x)\}}
    &\ge \mean{\x \sim \cd}{ \gamma(\x) \cdot \ind\{h(\x) \ne h^*(\x)\} \cdot \ind \{\gamma(\x) \ge \gamma_\delta(\cd)\}} \\
    &\ge \gamma_\delta(\cd)\P_\cd(A \cap B)
\end{split}
\end{equation}
where $A$ denotes the event where $h(\x) \ne h^*(\x)$ and $B$ denotes the event where $\gamma(\x) \ge \gamma_\delta(\cd)$. By definition of the loss we have $\P_\cd(A) = L_{\cd^*}(h)$, and by definition of the margin we have $\P_\cd(B) \ge 1-\delta$. Therefore, we have:
\begin{equation}
\label{eq:alb_3}
\P_\cd(A \cap B) = \P_\cd(A) + \P_\cd(B) - \P_\cd(A \cup B) \ge L_{\cd^*}(h) + (1-\delta) -1  = L_{\cd^*}(h) - \delta
\end{equation}

Now, combining Eq. (\ref{eq:alb_1}), (\ref{eq:alb_2}) and (\ref{eq:alb_3}), together with the fact that $L_\cd(h) \le L_\cd(h^*) + \epsilon$, we get:
\begin{align*}
    \gamma_\delta(\cd)(L_{\cd^*}(h) - \delta) + L_\cd(h^*) \le L_\cd(h) \le L_\cd(h^*)+\epsilon
\end{align*}
and so the required follows.

\end{proof}

\begin{proof}{of Theorem \ref{thm:erm_learner}.}

Fix some $\epsilon \in (0,1)$ and let $\epsilon' = \frac{\epsilon\gamma(\cd)}{2}$ and $\delta' = \frac{\epsilon}{2}$.
By the Fundamental Theorem of Statistical Learning (see \cite{shalev2014understanding}), there exists some universal constant $C$ s.t. taking $m = C \frac{\VC(\ch) + \log(1/\delta')}{(\epsilon')^2}$ we get that w.p. at least $1-\delta'$ over sampling $S \sim \cd^m$ it holds that:
\[
L_\cd(\ERM_\ch(S)) \le \inf_{h \in \ch}L_\cd(h) + \epsilon' = L_\cd(f^*_\cd) + \epsilon'
\]
where we use the fact that $f^*_\cd \in \ch$ is the Bayes optimal of $\cd$. Now, from Lemma \ref{lem:agnostic_loss_bound} it holds that, w.p. at least $1-\delta'$ it holds that (note that $\gamma(\cd)=\gamma_0(\cd)$),
\[
L_{\cd^*}(\ERM_\ch(S)) \le \frac{\epsilon'}{\gamma(\cd)}
\]
So, we get that:
\[
\E_{S \sim \cd^m} L_{\cd^*}(\ERM_\ch(S)) \le \frac{\epsilon'}{\gamma(\cd)} + \delta' = \epsilon
\]
\end{proof}

\begin{proof}{of Lemma \ref{lem:sampler-noise}.}

 Let the event $E=\{(\x,y)|\  y\ne f^*(\x)\}$, 
then,
\begin{align*}
    \left|\eta(\cd) - \E_{S \sim \cd^m} L_{\cd^*}(\ca(S))\right| &= \left|\P_{\x,y\sim\cd}[y\ne f^*(\x)] - \P_{\substack{\x\sim\cd_\cx\\ S\sim \cd^m}}[\ca(S)(\x)\ne f^*(\x)]\right| = \\
    &=\left|\cd(E) - \ca(\cd^m)(E)\right| \le \sup_{E} \left|\cd(E) - \ca(\cd^m)(E)\right|=\\
    & = TV(\cd, \ca(\cd^m)) = \eps  
\end{align*}\end{proof}

\begin{proof}{of Theorem \ref{thm:lower_bound}.}

We follow a proof similar to Chapter 28.2.1 of \citet{shalev2014understanding}.

Let $\gamma = \sqrt{\frac{\log(4/3)}{2M}}$. For every $b \in \{\pm 1\}$, let $\cd_b$ be the distributions concentrated on a single example $\x \in \cx$, with label, \[y \sim P_b(y) = \mathrm{Bernoulli}\inparen{\frac{1+b\gamma}{2}}= \begin{cases}  \frac{1+b\gamma}{2} & \mathrm{if}~y = 1 \\ \frac{1-b\gamma}{2} & \mathrm{if}~y = -1\end{cases}\]

Take $\cp = \{\cd_+, \cd_-\}$. Observe that the algorithm $\ca$ that takes a single sample $(\x,y_0)$ and outputs $y_0$ is a sampler for every $\cd \in \cp$.

Let $\y \in \{\pm 1\}^m$ be the sequence of labels observed by the algorithm $\ca$, and denote by $\ca(\y) \in \{ \pm 1\}$ the label that $\ca$ outputs for $\x$ when observing the sequence of labels $\y$. Note, that $\cd^*$ will be a \emph{constant distribution} concentrated on $(\x, b)$. Therefore, we have:
\begin{align*}
\E_{S \sim \cd_b^m} L_{\cd^*}(\ca(S)) = \E_{S \sim \cd_b^m} \ind \{\ca(S)(\x) \ne b\} = \E_{\y \sim P^m_b} \ind \{\ca(\y) \ne b\} 
\end{align*}
Denote $N_+ := \{\y \in \{\pm 1\}^m ~:~ \sum_i y_i \ge 0\}$ and $N_- = \{\pm 1\}^m \setminus N_+$. Then:
\begin{align*}
&\phantom{Q}\E_{\y \sim P_+^m}\ind\{\ca(\y) = -1\} + 
\E_{\y \sim P_-^m}\ind\{\ca(\y) = 1\} \\
&= \sum_{\y}P_+(\y) \ind \{\ca(\y) = -1\} + P_-(\y) \ind \{\ca(\y) = 1\} \\
&= \sum_{\y \in N_+}P_+(\y) \ind \{\ca(\y) =-1\} + P_-(\y) \ind \{\ca(\y) = 1\} \\
&+\sum_{\y \in N_-}P_+(\y) \ind \{\ca(\y) =-1\} + P_-(\y) \ind \{\ca(\y) = 1\} \\
& \ge \sum_{\y \in N_+}P_-(\y) + \sum_{\y \in N_-}P_+(\y) \ge \frac{1}{2} \inparen{1-\sqrt{1-\exp(-2m\gamma^2)}}
\end{align*}
where the last inequality follows from Lemma B.11 in \cite{shalev2014understanding}.
So, if $m < \frac{\log(4/3)}{2\gamma^2} = M$ we get:
\[
\E_b \E_{S \sim \cd_b^m} L_{\cd^*}(\ca(S)) = \frac{1}{2} \inparen{\E_{\y \sim P_+^m}\ind\{\ca(\y) = -1\} + 
\E_{\y \sim P_-^m}\ind\{\ca(\y) = 1\}}> \frac{1}{8}
\]
and we get there exists $\cd \in \cp$ s.t. if $m < M$ then $\E_{S \sim \cd^m} L_{\cd^*}(\ca(S)) > \frac{1}{8}$.
\end{proof}

\begin{proof}{of Theorem \ref{thm:sampler_teacher}.}

To see property \textit{1.} of Definition~\ref{def:teacher}, we show that if two distributions over $(\x, y)$ are close in total variation, then the Bayes optimal classifier for both has to be similar. That is, $$\mathrm{TV}(\cd, \cd')<\eps \quad \Longrightarrow \quad \prob{\x \sim \cd}{f^*_\cd(\x) \ne f^*_{\cd'}(\x)}\le \eps/\gamma$$
Note, for $\x\sim\cd_\cx$  we have $y_\x =: f^*_\cd(\x) \ne f^*_{\cd'}(\x) =: \hat y_\x$ if and only if $\prob{\D'}{y_\x|\x} < \prob{\D'}{\hat y_\x| \x}$, but the margin condition guarantees that $\prob{\D}{y_\x|\x} - \prob{\D}{\hat y_\x| \x} \ge \gamma$, thus,
\begin{align*}
    \PP_{\x \sim \cd}[f^*_\cd(\x) \ne f^*_{\cd'}(\x)]
    &\le\P_{\x\sim\cd_\cx}[|\prob{\D}{y_\x|\x} - \prob{\D'}{y_\x|\x}| > \gamma] \le\\
    &\le \E_{\x \sim \cd}[\left|\prob{\D}{y_\x|\x} - \prob{\D'}{y_\x|\x}\right|]/\gamma.
\end{align*}
Where we use Markov inequality for the second transition. Now, we can use the alternative definition of TV to conclude the proof (here $\p(\x)$ is the Radon–Nikodym measure  of $\x$ under the marginal $\cd_\cx$ and $\p_\cd(\x, y)$ is the Radon–Nikodym measure of $(\x, y)$ under $\cd$):
\begin{align*}
    \E_{\x\sim\cd_\cx}[\left|\prob{\D}{y_\x|\x}-\prob{\D'}{y_\x|\x}\right|] &= \int \left|\prob{\D}{y_\x|\x}-\prob{\D'}{y_\x|\x}\right| \p(\x) \le \\
    &\le \frac{1}{2}\int \left|\p_\cd(\x, y)-\p_{\cd'}(\x, y)\right| \le  \eps
\end{align*}
Where the penultimate inequality is based on an easy corollary of the triangle inequality: $
\forall y$ we have $$\sum _y |\PP_\cd[y|\x] - \PP_{\cd'}[y|\x] | \ge  2|\PP_\cd[y|\x] - \PP_{\cd'}[y|\x] |.$$

We proceed to prove the 2$^{\text{rd}}$ property. For each $\x\in\cx$ let $y_1$ and $y_2$ be the two most likely labels respectively with respect to the distribution  $\cd$, that is, $y_1(\x) = \arg\max_y \P_\cd[y| \x]$ and $y_2(\x) = \arg\max_{y\ne y_1(\x)} \P_\cd[y| \x]$ and $y_1',y_2'$ defined similarly for $\cd'$. Then, for a given $\x$, if the margin is small, i.e., $\prob{\cd'}{y_1'|\x}-\prob{\cd'}{y_2'|\x}<\gamma-\tau$ then we will want to prove that the following holds:
\begin{align}
\sum_y |\prob{\cd}{y|\x} - \prob{\cd'}{y|\x}| &> \tau \label{eq:sum_large}
\end{align}

\noindent If $y_1 \ne y_1'$ then  with probability 1 we have $\prob{\cd}{y_1|\x} - \prob{\cd}{y_1'|\x}> \gamma$. Using the definition  of $y_1'$,
$$\prob{\cd}{y_1|\x} - \prob{\cd}{y_1'|\x} + \prob{\cd'}{y'_1|\x} - \prob{\cd'}{y_1|\x} > \gamma > \tau$$

\noindent If, on the other hand, $y_1 = y_1'$ using $\prob{\cd}{y_1|\x}-\prob{\cd}{y_2|\x}>\gamma$ again we have (by summing up the inequalities):
\begin{align*}
\prob{\cd'}{y_1'|\x}-\prob{\cd'}{y_2'|\x} + \prob{\cd'}{y_2'|\x}-\prob{\cd'}{y_1'|\x} &> \tau 
\end{align*}
So in both cases Equation~\ref{eq:sum_large} holds. Thus, 

\begin{align*}
\P_{\x}\left[\prob{\cd'}{y_1'(\x)|\x} - \prob{\cd'}{y_2'(\x)|\x}<\gamma-\tau\right] & \le \\
\P_{\x}\left[
\sum_y |\prob{\cd}{y|\x} - \prob{\cd'}{y|\x}| > \tau
\right] & \le \\
\E _{\x}\left[
\sum_y |\prob{\cd}{y|\x} - \prob{\cd'}{y|\x}| 
\right] / \tau & = 2\mathrm{TV}(\cd, \cd')/\tau = 2\frac\eps\tau
\end{align*}

\end{proof}

\begin{proof}{of Theorem \ref{thm:learner_teacher}.}

Fix $\epsilon \in (0,1)$ and $0 < \tau < \gamma(\cd)$.
Let $m = m\left(\frac{\epsilon(1-\gamma(\cd)+\tau)}{2}\right)$. Fix some $\x \in \cx$ such that \[
f^*_\cd(\x) \ne f^*_{\ca(\cd^m)}(\x) = \arg \max_y \prob{\ca(\cd^m)}{y|\x}
\]
Then,
\[
\prob{S \sim \cd^m}{\ca(S)(\x) \ne f^*_\cd(\x)} = \prob{(\x,y)\sim\ca(\cd^m)}{y \ne f^*_\cd(\x)|\x} \ge \frac{1}{2}
\]
Therefore, since $\ca$ is a learner with sample complexity $m(\cdot)$ we have:
\begin{align*}
    \frac{\epsilon}{2} &\ge \E_{S \sim \cd^m} L_{\cd^*}\left(\ca(S)\right) = \E_{\x\sim\cd_\cx}\PP_{S \sim \cd^m}\left[\ca(S)(\x) \ne f^*_\cd(\x)\right] \\
    &\ge \mean{\x}{\PP_{S}{[\ca(S)(\x) \ne f^*_\cd(\x)]} \middle| f^*_\cd(\x) \ne f^*_{\ca(\cd^m)}(\x)}\cdot \prob{\x}{f^*_\cd(\x) \ne f^*_{\ca(\cd^m)}(\x)} \\
    & \ge \frac{1}{2}\prob{\x}{f^*_\cd(\x) \ne f^*_{\ca(\cd^m)}(\x)} = \frac{1}{2}L_{\cd^*}\left(f^*_{\ca(\cd^m)}\right)
\end{align*}
So, the first condition of Definition \ref{def:teacher} holds. For the second condition, observe that since $f^*_{\ca(\cd^m)}$ is the Bayes optimal classifier, we have:
\begin{align*}
    \eta(\ca(\cd^m)) &= \prob{(\x,y) \sim \ca(\cd^m)}{y \ne f^*_{\ca(\cd^m)}(\x)} \le \prob{\ca(\cd^m)}{y \ne f^*_{\cd}(\x)} \\
    &= \E_{\x \sim \cd} \PP_{S \sim \cd^m}\left[\ca(S)(\x) \ne f^*_\cd(\x)\right]
    = \E_{S \sim \cd^m} L_{\cd^*}(\ca(S)) \le  \frac{\delta(1-\gamma(\cd)+\tau)}{2}
\end{align*}
where the last inequality is using the fact that $\ca$ is a learner.
From Lemma \ref{lem:noise-margin}, since $\eta(\ca(\cd)) \le \frac{\delta(1-\gamma(\cd)+\tau)}{2}$, it holds that $\gamma_{\delta}(\ca(\cd)) \ge \gamma(\cd) -\tau$.
\end{proof}

\begin{lemma}
\label{lem:noise-margin}
Let $\cd$ be a distribution with $\mu$-bounded  noise, i.e., $\eta(\cd) = \PP_{\x,y\sim \cd}[y\ne f^*(\x)]\le  \mu$ where $f^*$ is the Bayes optimal classifier. Let $0<\gamma<1$ be some positive constant denoting a margin. Then, \[
\PP_{\x\sim\cd_\cx}\left[\P_{\cd}(f^*_\cd(\x)|\x) < \max_{y \ne f^*_\cd(\x)} \P_{\cd}(y|\x) + \gamma \right] \le \frac{2\cdot\mu} {(1-\gamma)}
\]
\end{lemma}
\begin{proof}
For each $\x\in\cx$ let $y_1(\x)$ and $y_2(\x)$ be the two most likely labels respectively, that is, $y_1(\x) = \arg\max_y \P[y| \x]$ and $y_2(\x) = \arg\max_{y\ne y_1(\x)} \P[y| \x]$. Let $\gamma_\x = \P[y_1(\x)|\x]-\P[y_2(\x)|\x]$ and denote the set of small margin examples  $B=\{\x| \gamma_\x \le \gamma\}$. Then we have,
\begin{align*}
    \eta(\cd) &=  \E_{\x} [\P[Y\ne y_1(\x) | \x]] \ge \\
    & \ge \E_{\x}[\P[Y\ne y_1(\x) | \x]|\x \in B] \P[\x \in B] \ge \\
    & \ge \P(B) \cdot \frac{1-\gamma}{2}
\end{align*}
Where the last inequality is proven via the following lemma:
\end{proof}

\begin{lemma}
Given a fixed $\x\in B$ s.t. $\gamma_\x \le \gamma$ (in notation of Lemma~\ref{lem:noise-margin}) for some $0 < \gamma < 1$. Then, 
\[
\P[Y\ne y_1(\x) | \x] \ge \frac{1-\gamma}{2}
\]
\end{lemma}
\begin{proof}
Since the $\x$ is fixed we drop all $\x$ related notation WLOG:
\begin{align*}
    \P[Y = y_1] & = 1-\P[Y\ne y_1] \le \\
    & \le 1-\P[Y=y_2]\le \\
    & \le 1 - \P[Y=y_1] + \gamma
\end{align*}
Thus, by rearranging we get $\P[Y = y_1] \le \frac{1+\gamma}{2}$ which implies $\P[Y \ne y_1] \ge \frac{1-\gamma}{2}$
\end{proof}

\section{Proofs of Section \ref{sec:distillation}}

To prove Theorem \ref{thm:ensemble_inference} we use the following Lemma:
\begin{lemma}
\label{lem:point_bound}
Let $\ca$ be some learning algorithm. Fix some $\gamma > 0$ and $\tau < \gamma$. Then, for every $\x$ s.t.
\begin{itemize}
    \item $\prob{\ca(\cd^m)}{f^*_{\ca(\cd^m)}(\x)\mid\x} > \prob{\ca(\cd^m)}{-f^*_{\ca(\cd^m)}(\x)\mid\x}+\gamma$ and
    \item $f^*_{\ca(\cd^m)}(\x) = f^*_\cd(\x)$
\end{itemize}
it holds that: $$\prob{S_1, \dots, S_k \sim \cd^m}{f_\cd^*(\x)\frac{1}{k}\sum_{i=1}^k \ca(S_i)(\x) \le \tau } \le \exp \inparen{-\frac{k (\gamma-\tau)^2}{4}}$$
\end{lemma}

\begin{proof}{of Lemma \ref{lem:point_bound}.}

Fix some $\x \in \cx$ and denote
$$p_\x(y) = \prob{\ca(\cd^m)}{y | \x} = \prob{S \sim \cd^m}{\ca(S)(\x) = y}$$
Let $y^*_\x = \arg \max_y p_\x(y) = f^*_{\ca(\cd^m)}(\x)$.
So, assume that $\x$ satisfies the assumption, namely assume that $p_\x(y_\x^*) > p_\x(-y_\x^*) + \gamma$ and $y^*_\x = f^*_\cd(\x)$.

Denote $y_\x^{(i)} = \ca(S_i)(\x)$, the prediction of the $i$-th teacher on $\x$. Then, 
$$\mean{}{\frac{1}{k}y^*_\x\sum_{i=1}^k y_\x^{(i)}} = \mean{}{y_\x^* y_\x^{(1)}} =  p_\x(y^*_\x)-p_\x(-y^*_\x)> \gamma$$
By Hoeffding's inequality we get:
$$\prob{}{\frac{1}{k}y^*_\x\sum_{i=1}^k y_\x^{(i)}\le \tau } \le \exp \inparen{-\frac{k\inparen{\mean{}{\frac{1}{k}y_\x^* \sum_{i=1}^k y_\x^{(i)}}-\tau}^2}{4}} = \exp \inparen{-\frac{k (\gamma-\tau)^2}{4}}$$
\end{proof}

\begin{proof}{of Theorem \ref{thm:ensemble_inference}}

Let $\cx' \subseteq \cx$ be the subset of points $\x \in \cx$ satisfying the assumptions of Lemma \ref{lem:point_bound} with $\gamma = \frac{\gamma(\cd)}{2}$ and $\tau = 0$. Observe that, using the union bound, and the properties of the teacher $\ca$:
\begin{align*}
&\prob{\x \sim \cd}{\x \notin \cx'} \\
&\le \prob{\x \sim \cd}{\prob{\ca(\cd^m)}{f^*_{\ca(\cd^m)}(\x)\mid\x} > \prob{\ca(\cd^m)}{-f^*_{\ca(\cd^m)}(\x)\mid\x}+\gamma} \\
&+ \prob{\x \sim \cd}{f^*_{\ca(\cd^m)}(\x) \ne f^*_\cd(\x)} \\
&\le \epsilon/3 + L_{\cd^*}\inparen{f^*_{\ca(\cd^m)}} \le \frac{2\epsilon}{3}
\end{align*}

Now, fix some $\x \in \cx'$, and from Lemma \ref{lem:point_bound} we have:
\[
\E_{S_1, \dots, S_k \sim \cd^m} \ind\{\ca_\ens(S_1, \dots, S_k)(\x) \ne f_\cd^*(\x)\} \le \exp\inparen{-\frac{k\gamma^2}{4}} \le \epsilon/3
\]

Finally, we get:
\begin{align*}
&\E_{S_1, \dots, S_k \sim \cd^m} L_{\cd^*}(\ca_\ens(S_1, \dots, S_k)) \\
&= \E_{S_1, \dots, S_k \sim \cd^m} \E_\x \ind\{\ca_\ens(S_1, \dots, S_k)(\x) \ne f^*_\cd(\x)\} \\
&= \prob{\x \sim \cd}{\x \in \cx'} \cdot \E_{\x | \x \in \cx'}\E_{S_1, \dots, S_k \sim \cd^m}  \ind\{\ca_\ens(S_1, \dots, S_k)(\x) \ne f^*_\cd(\x)\} \\
&+\prob{\x \sim \cd}{\x \notin \cx'} \cdot \E_{\x | \x \notin \cx'}\E_{S_1, \dots, S_k \sim \cd^m}  \ind\{\ca_\ens(S_1, \dots, S_k)(\x) \ne f^*_\cd(\x)\} \\
&\le \inparen{1-\frac{2\epsilon}{3}}\frac{\epsilon}{3}+\frac{2\epsilon}{3} \le \epsilon
\end{align*}

\end{proof}

\begin{proof}{of Theorem \ref{thm:ensemble_pseudo_labeling}.}

Fix a sequence of $k$ subsets of examples $\cs = (S_1, \dots, S_k)$, and let $\widetilde{\cd}_\cs$ be the distribution given by sampling $\x \sim \cd$ and returning $(\x, y)$ where $y = \ca_\ens(S_1, \dots, S_k)(\x)$. Let $\tilde{S}_\cs$ be an i.i.d. sample of size $m'$ from $\widetilde{\cd}_\cs$. Let $h_{\cs} = \ERM_\ch(\widetilde{S}_\cs)$. By the Fundamental Theorem of Statistical Learning (e.g. Theorem 6.8 in \cite{shalev2014understanding}) w.p. at least $1-\epsilon/4$ over sampling $\widetilde{S}_\cs$ we have:
\begin{align*}
L_{\widetilde{\cd}_\cs}(h_{\cs}) &\le \inf_{h \in \ch} L_{\widetilde{\cd}_\cs}(h) + \epsilon/4
\le L_{\widetilde{\cd}_\cs}(f^*_\cd)+\epsilon/4 \\
&= \prob{\x \sim \cd}{\ca_\ens(\cs)(\x) \ne f^*_\cd(\x)} + \epsilon/4 = L_{\cd^*}(\ca_\ens(\cs)) + \epsilon/4
\end{align*}
On the other hand, observe that for all $h$:
\begin{align*}
L_{\cd^*}(h) &= \E_{\x \sim \cd} \ind \{h(\x) \ne f^*_\cd(\x)\} \\
&\le \E_{\x \sim \cd} \inparen{\ind \{h(\x) \ne \ca_\ens(\cs)(\x)\} + \ind \{\ca_\ens(\cs)(\x) \ne f^*_\cd(\x)\}} \\
&= L_{\tcd_\cs}(h) + L_{\cd^*}(\ca_\ens(\cs))
\end{align*}
Overall we get that w.p. at least $1-\epsilon/4$ over sampling $\widetilde{S}_\cs$ we have:
\[
L_{\cd^*}(h_\cs) \le 2 L_{\cd^*}(\ca_\ens(\cs)) + \epsilon/4
\]
and therefore:
\[
\E_{\widetilde{\cs}_\cs \sim \tcd^{m'}_\cs} L_{\cd^*}(h_\cs) \le 2L_{\cd^*}(\ca_\ens(\cs)) + \epsilon/2
\]
Finally, using Theorem \ref{thm:ensemble_inference} we get:
\[
\E_{S_1, \dots, S_k, \widetilde{S}} L_{\cd^*}(h) \le 2\E_{S_1, \dots, S_k \sim \cd^m} L_{\cd^*}(\ca_\ens(S_1, \dots, S_k)) + \epsilon/2 \le \epsilon
\]
where $h$ is the output of the Ensemble-Pseudo-Labeling algorithm.
\end{proof}

To prove Theorem \ref{thm:random_ensemble}, we use the following Lemma:

\begin{lemma}
\label{lem:ensemble_confidence}
Assume that $\ca$ is a \textbf{teacher} for some distribution $\cd$, with sample complexity $\widetilde{m}$. Then, for every $\epsilon, \delta \in (0,1)$, taking $m \ge \widetilde{m}\inparen{\frac{\epsilon}{3}, \frac{\gamma(\cd)}{2}}$ and $k \ge 
\frac{64}{\gamma(\cd)^2}\log\inparen{\frac{3}{\epsilon\delta}}$ we get that w.p. at least $1-\delta$ over the choice of $S_1, \dots, S_k$, it holds that:
\[
\prob{\x \sim \cd}{f^*_\cd(\x)\frac{1}{k}\sum_{i=1}^k \ca(S_i)(\x) \le \gamma(\cd)/4} \le \epsilon
\]
\end{lemma}

\begin{proof}{of Lemma \ref{lem:ensemble_confidence}.}
Let $\cx' \subseteq \cx$ be the subset of points $\x \in \cx$ satisfying the assumptions of Lemma \ref{lem:point_bound} with $\gamma = \frac{\gamma(\cd)}{2}$ and $\tau = \frac{\gamma(\cd)}{4}$. Observe that, using the union bound, and the properties of the teacher $\ca$:
\begin{align*}
&\prob{\x \sim \cd}{\x \notin \cx'} \\
&\le \prob{\x \sim \cd}{\prob{\ca(\cd^m)}{f^*_{\ca(\cd^m)}(\x)\mid\x} > \prob{\ca(\cd^m)}{-f^*_{\ca(\cd^m)}(\x)\mid\x}+\gamma} \\
&+ \prob{\x \sim \cd}{f^*_{\ca(\cd^m)}(\x) \ne f^*_\cd(\x)} \\
&\le \epsilon/3 + L_{\cd^*}\inparen{f^*_{\ca(\cd^m)}} \le \frac{2\epsilon}{3}
\end{align*}

Let $\delta' = \frac{\epsilon\delta}{3}$. Fix some $\x \in \cx'$, and from Lemma \ref{lem:point_bound} we have:
\[
\E_{S_1, \dots, S_k \sim \cd^m} \ind\{f^*_\cd(\x)\frac{1}{k}\sum_i\ca_\ens(S_i)(\x) \le \tau \} \le \exp\inparen{-\frac{k(\gamma-\tau)^2}{4}} \le \delta'
\]

Therefore, we get:
\begin{align*}
&\E_{S_1, \dots, S_k \sim \cd^m} \P_{\x \sim \cd} \left[f^*_\cd(\x)\frac{1}{k}\sum_i\ca_\ens(S_i)(\x) \le \tau \mid \x \in \cx'\right] \\
&= \E_\x \left[\E_{S_1, \dots, S_k \sim \cd^m} \ind\{f^*_\cd(\x)\frac{1}{k}\sum_i\ca_\ens(S_i)(\x) \le \tau\}  \mid \x \in \cx' \right] \le \delta'
\end{align*}

Using Markov's inequality we get that w.p. at least $1-\frac{3 \delta'}{\epsilon}$ we have $$\P_{\x \sim \cd}\left[ f^*_\cd(\x)\frac{1}{k}\sum_i\ca_\ens(S_i)(\x) \le \tau \mid \x \in \cx'\right] \le \frac{\epsilon}{3}$$
and in this case we have
\begin{align*}
&\P_{\x \sim \cd} \left[f^*_\cd(\x)\frac{1}{k}\sum_i\ca_\ens(S_i)(\x) \le \tau \right] \\
&\le \P_{\x \sim \cd}\left[ f^*_\cd(\x)\frac{1}{k}\sum_i\ca_\ens(S_i)(\x) \le \tau \mid \x \in \cx'\right] + \P_{\x \sim \cd} \left[\x \notin \cx'\right] \le \epsilon
\end{align*}
\end{proof}

\begin{proof}{of Theorem \ref{thm:random_ensemble}.}
Fix $\epsilon > 0$ and let $\epsilon' = \frac{\gamma(\cd)\epsilon}{18}$.
Fix a sequence of $k$ subsets of examples $\cs = (S_1, \dots, S_k)$, and let $\widetilde{\cd}_\cs$ be the distribution over $\cx \times \cy$ given by sampling $\x \sim \cd$, sampling $i \sim \{1, \dots, k\}$ and returning $(\x, y)$ where $y=\ca(S_i)(\x)$. Let $\widetilde{S}_\cs$ be an i.i.d. sample of size $m'$ from $\widetilde{\cd}_\cs$. Let $h_{\cs} = \ERM_\ch(\widetilde{S}_\cs)$. By the Fundamental Theorem of Statistical Learning (e.g. Theorem 6.8 in \cite{shalev2014understanding}) w.p. at least $1-\epsilon'$ over sampling $\widetilde{S}_\cs$ we have:
\begin{align*}
L_{\tcd_\cs}(h_{\cs}) &\le \inf_{h \in \ch} L_{\tcd_\cs}(h) + \epsilon' \le L_{\tcd_\cs}(f^*_\cd)+\epsilon' \\
&= \E_{x \sim \cd}[\ind\{f^*_\cd(\x) \ne y\}]
\le \E_{x \sim \cd}[\ind\{f^*_\cd(\x) \ne \ca_\ens(\cs)(\x)\} + \ind\{\ca_\ens(\cs)(\x) \ne y\}] +\epsilon'\\
&= L_{\cd^*}(\ca_\ens(\cs)) + L_{\tcd_\cs}(\ca_\ens(\cs)) + \epsilon'
\end{align*}

\noindent\textbf{Claim}: If $\cs$ satisfies $\gamma_{\epsilon'}(\tcd_\cs) > 0$ then w.p. at least $1-\epsilon'$ over the choice of $\widetilde{S}_\cs \sim \tcd_\cs^{m'}$
$$L_{\cd^*}(h_\cs) \le (L_{\cd^*}(\ca_\ens(\cs)) + \epsilon')\inparen{1+\gamma_{\epsilon'}(\tcd_\cs)^{-1}}$$

\noindent\textbf{Proof}: W.p. at least $1-\epsilon'$ we have $L_{\tcd_\cs}(h_\cs) \le L_{\tcd_\cs}(\ca_\ens(\cs)) + L_{\cd^*}(\ca_\ens(\cs)) + \epsilon'$. 
Notice that by definition of $\tcd_\cs$, we have that $\ca_\ens(\cs)$ is the Bayes optimal classifier for $\tcd_\cs$. Therefore, by Lemma \ref{lem:agnostic_loss_bound} we have $L_{\tcd_\cs^*}(h_\cs) \le \frac{\epsilon'+L_{\cd^*}(\ca_\ens(\cs))}{\gamma_{\epsilon'}(\tcd_\cs)} + \epsilon'$. Now, we have:
\begin{align*}
L_{\cd^*}(h_\cs) &= \E_{\x \sim \cd}[\ind\{h_\cs(\x) \ne f^*_\cd(\x)\}] \\
&\le \E_{\x \sim \cd}[\ind\{h_\cs(\x) \ne \ca_\ens(\cs)(\x)\} + \ind\{\ca_\ens(\cs)(\x) \ne f^*_\cd(\x)\}] \\
&= L_{\tcd_\cs^*}(h_\cs) + L_{\cd^*}(\ca_\ens(\cs)) \le \frac{\epsilon'+L_{\cd^*}(\ca_\ens(\cs))}{\gamma_{\epsilon'}(\tcd_\cs)} + \epsilon' + L_{\cd^*}(\ca_\ens(\cs))
\end{align*}

\noindent\textbf{Claim}: W.p. $>1-\epsilon'$ over the choice of $\cs$, we have $\gamma_{\epsilon'}(\tcd_\cs) \ge \frac{\gamma(\cd)}{4}$ and $L_{\cd^*}(\ca_\ens(\cs)) \le \epsilon'$.

\noindent\textbf{Proof}: By Lemma \ref{lem:ensemble_confidence}, since $m \ge \widetilde{m}\inparen{\frac{\epsilon'}{3}, \frac{\gamma(\cd)}{2}, \frac{\epsilon'}{3}}$ and $k \ge \frac{64}{\gamma(\cd)^2}\log\inparen{\frac{3}{(\epsilon')^2}}$ we have, w.p. $> 1-\epsilon'$ over the choice of $\cs$, that
\begin{align*}
&\P_{\x \sim \cd}\left[(\P_{i \sim [k]}[\ca(S_i)(\x) = f^*_\cd(\x)|\x]-\P_{i \sim [k]}[\ca(S_i)(\x) = -f^*_\cd(\x)|\x]) > \gamma(\cd)/4\right] \\
&=\P_{\x \sim \cd} \left[f^*_\cd(\x) \frac{1}{k}\sum_i \ca(S_i)(\x) > \gamma(\cd)/4\right] \le \epsilon'
\end{align*}
which immediately implies the required.

From the above two claims, w.p. at least $1-2\epsilon'$ over the choice of $\cs, \widetilde{S}_\cs$ we have
$$L_{\cd^*}(h_\cs) \le 2\epsilon'\inparen{1+\gamma_{\epsilon'}(\tcd_\cs)^{-1}}\le \frac{16 \epsilon'}{\gamma(\cd)}$$
and therefore $\E_{\cs,\widetilde{S}_{\cs}}L_{\cd^*}(h_\cs) \le \frac{16\epsilon'}{\gamma(\cd)} + 2\epsilon' \le \frac{18\epsilon'}{\gamma(\cd)} = \epsilon$.
\end{proof}

\section{Proofs of Section \ref{sec:algorithms}}
Using standard measure-theoretic arguments, we show that for any distribution $\cd$, such a cover exists: 

\begin{lemma}
\label{lem:covering}
For a every distribution $\cd$ over $\cx \times \cy$ there exists a function $m_c : (0,1) \times (0,1) \to \naturals$ s.t. for every $\epsilon,\delta \in (0,1)$ there exists a subset $\cx' \subseteq \cx$ satisfying:
\begin{itemize}
    \item $\prob{\x \sim \cd_\cx}{\x \notin \cx'} \le \delta$
    \item If $m \ge m_c(\epsilon,\delta)$, for all $\x \in \cx'$ it holds that $\prob{S \sim \cd_\cx^m}{d(\x,S) > \epsilon} \le \delta$.
\end{itemize}
\end{lemma}
\begin{proof}{of Lemma \ref{lem:covering}}
Fix some $\epsilon, \delta \in (0,1)$ and let $\delta' = \delta/2, \epsilon' = \epsilon/2$. For some $\x_0 \in \cx$ and let $B_r(\x_0)$ be the closed ball of radius $r$ around $\x_0$, i.e. $$B_r(\x_0) = \{\x \in \cx ~:~ d(\x_0,\x) \le r\}$$
Now, for some $\x_0 \in \cx$, observe that $\cx = \cup_{r=1}^\infty B_r(\x_0)$, and therefore we have:
\[
1 = \cd_\cx(\cx) = \cd_\cx(\cup_{r=1}^\infty B_r(\x_0)) = \lim_{r \to \infty} \cd_\cx(B_r(\x_0))
\]
So, there exists some $r$ s.t. $\cd_\cx(B_r(\x_0)) \ge 1-\delta'$. Now, since $B_r(\x_0)$ is closed and bounded in $(\cx,d)$, from the Heine-Borel property we get that $B_r(\x_0)$ is also compact. Since $B_r(\x_0) \subseteq \cup_{\x \in B_r(\x_0)} B_{\epsilon'}(\x)$, there exists some finite subset $C \subseteq B_r(\x_0)$ such that $B_r(\x_0) \subseteq \cup_{\x \in C} B_{\epsilon'}(\x)$. 
Now, let $C' \subseteq C$ be the subset of balls that have at least $\delta'/\abs{C}$ mass under $\cd_\cx$, namely:
\[
C' = \inbrace{\x \in C ~:~ \cd_\cx(B_{\epsilon'}(\x)) \ge \frac{\delta'}{\abs{C}}}
\]
Let $m = \left\lceil \frac{\abs{C}}{\delta'} \log \inparen{\frac{\abs{C}}{\delta'}} \right\rceil$, and observe that for every $\x \in C'$ we have:
\[
\prob{S \sim \cd_\cx^m}{S \cap B_{\epsilon'}(\x) = \emptyset} = \prob{\x' \sim \cd_\cx}{\x' \notin B_{\epsilon'}(\x)}^m \le \inparen{1-\frac{\delta'}{\abs{C}}}^m \le \exp\inparen{-\frac{m\delta'}{\abs{C}}} \le \frac{\delta'}{\abs{C}}
\]
Using the union bound, w.p. at least $1-\delta'$ it holds that for all $\x \in C'$ ther exists $\x' \in S$ s.t. $\x' \in B_{\epsilon'}(\x)$. Denote by $\cx'$ all the points in $\cx$ that are covered by $C'$, namely
$\cx' = \cup_{\x \in C'} B_{\epsilon'}(\x)$.

\textbf{Claim}: $\cx \setminus \cx' \subseteq (\cx \setminus B_r(\x_0)) \cup (\cup_{\x \in C \setminus C'} B_{\epsilon'}(\x))$

\textbf{Proof}: Let $\x \in \cx \setminus \cx'$ and we need to show $\x \in (\cx \setminus B_r(\x_0)) \cup (\cup_{\x \in C \setminus C'} B_{\epsilon'}(\x))$. If $\x \notin B_r(\x_0)$ we are done. Otherwise, if $\x \in B_r(\x_0)$, since $B_r(\x_0) \subseteq \cup_{\x' \in C} B_{\epsilon'}(\x)$ there exists some $\x' \in C$ s.t. $\x \in B_{\epsilon'}(\x')$, and $\x' \notin C'$ since otherwise we would have $\x \in \cx'$.

\textbf{Claim}:
$\prob{\x \sim \cd_\cx}{\x \notin \cx'} \le 2 \delta'$

\textbf{Proof}: Using the union bound and the previous result:
\begin{align*}
\prob{\x \sim \cd_\cx}{\x \notin \cx'}
&\le \prob{\x \sim \cd_\cx}{\x \notin B_r(\x_0)} + \sum_{\x' \in C \setminus C'} \prob{\x \sim \cd_\cx}{\x \in B_{\epsilon'}(\x')} \\
&\le \delta' + \abs{C\setminus C'} \frac{\delta'}{\abs{C}} \le 2 \delta'
\end{align*}

\textbf{Claim}: W.p. at least $1-\delta'$ over the choice of $S \sim \cd_\cx^m$, for all $\x \in \cx'$ it holds that $d(\x,S) \le \epsilon$.

\textbf{Proof}: From what we showed, w.p. at least $1-\delta'$, for all $\x \in C'$ there exits $\x' \in S$ s.t. $\x' \in B_{\epsilon'}(\x)$. Assume this holds, and let $\x \in \cx'$. By definition of $\cx'$ there exists some $\hat{\x} \in C'$ s.t. $d(\x, \hat{\x}) \le \epsilon'$. So, there is some $\x' \in S$ s.t. $d(\x',\hat{\x}) \le \epsilon'$, and therefore $d(\x, \x') \le 2 \epsilon' = \epsilon$ and we get the required.

Now, the required follows from the last two claims.
\end{proof}

\begin{proof}{of Theorem \ref{thm:one_nearest_neighbour}.}

Let $\cd$ be some $\lambda$-Lipschitz distribution. Let $m_c(\cdot, \cdot)$ be a function satisfying the conditions guaranteed by Lemma \ref{lem:covering} for the distribution $\cd$. Then, we prove that the $\ca_{\oNN}$ is a \textbf{sampler} for $\cd$, with distributional sample complexity $\widetilde{m}(\epsilon) = m_c\inparen{\frac{\epsilon}{2\lambda}, \frac{\epsilon}{12}}$.

Fix $\epsilon \in (0,1)$ and let $\epsilon' = \frac{\epsilon}{2\lambda}, \delta' = \frac{\epsilon}{12}$.
Let $m_c$ be the function guaranteed by Lemma \ref{lem:covering}, and let $\cx'$ be the subset guaranteed by the same Theorem (given the choice of $\epsilon', \delta'$).
Fix some $\x \in \cx'$. Denote $q := \prob{S \sim \cd^m}{d(\x,S) \le \epsilon'}$ (the probability to get a good cover). By Lemma \ref{lem:covering}, for $m = m_c(\epsilon',\delta')$ we get that $q \ge 1-\delta'$. For every $y \in \cy$, denote $p_\x(y) := \P_\cd [y|\x]$, and we have:
\begin{align*}
\abs{\P_{S \sim \cd^m} [\ca_{\NN}(S)(\x) = y ] - p_\x(y)} 
&\le q\abs{\P_{S \sim \cd^m} [\ca_{\NN}(S)(\x) = y| d(\x,S) \le \epsilon]-p_\x(y)} \\
&+(1-q)\abs{\P_{S \sim \cd^m} [\ca_{\NN}(S)(\x) = y| d(\x,S) > \epsilon]- p_\x(y)} \\
&\le \abs{\P_{\cd}[y|\pi(\x,S), d(\x,S) \le \epsilon]-p_\x(y)} + 2 \delta' \le \lambda \epsilon' + 2 \delta'
\end{align*}
From the above we get:
\begin{align*}
&\E_{\x}{\sum_{y} \abs{\P_{S \sim \cd^m}[\ca_\NN(S)(\x) | \x] - \P[y|\x]}} \\
&\le \E_{\x|\x \in \cx'} \sum_y \abs{\P_{S \sim \cd^m}[\ca_\NN(S)(\x) | \x] - \P[y|\x]} +2 \abs{\cy}\P_{\x \sim \cd}[\x \notin \cx] \\
&\le \lambda \epsilon' + 6\delta' \le \epsilon
\end{align*}
and therefore the required follows.
\end{proof}

\begin{proof}{of Theorem \ref{thm:knn}}

Let $\cd$ be some $\lambda$-Lipschitz distribution. Let $m_c(\cdot, \cdot)$ be a function satisfying the conditions guaranteed by Lemma \ref{lem:covering} for the distribution $\cd$. Then, we prove that the $\ca_{\kNN}$ algorithm is a \textbf{teacher} for $\cd$, with sample complexity $\widetilde{m}(\epsilon, \tau) = k \cdot m_c\inparen{\frac{\tau}{4\lambda},\min \inbrace{\epsilon, \frac{\tau}{4k}}}$.

Fix $\epsilon \in (0,1)$, $\tau \in (0,\gamma(\cd))$ and let $\epsilon' = \frac{\tau}{4\lambda}, \delta' = \min \inbrace{\epsilon, \frac{\tau}{4k}}$.
Let $m_c$ be the function guaranteed by Lemma \ref{lem:covering}, and let $\cx'$ be the subset guaranteed by the same Theorem (given the choice of $\epsilon', \delta'$).
Fix some $\x \in \cx'$. Let $\cs$ be the set of subsets of $\cx$ such that $S_\cx \in \cs$ if and only if for all $\x' \in \kpi(\x,S_\cx)$ it holds that $d(\x,\x') \le \epsilon'$. Let $m = k \cdot m_c(\epsilon', \delta')$.

\noindent\textbf{Claim:} $\prob{S \sim \cd^m}{S_\cx \in \cs} \ge 1-k\delta'$

\noindent\textbf{Proof:} For every set $S \subseteq \cx \times \cy$ of size $m$, split $S$ to blocks of $k$ examples $S^{(1)}, \dots, S^{(k)}$ each of size $m_c(\epsilon', \delta')$. By Lemma \ref{lem:covering}, for every $i$ it holds that $\prob{S^{(i)} \sim \cd^{m/k}}{d\inparen{\x,S^{(i)}_\cx} > \epsilon'} \le \delta'$. Using the union bound, with probability at least $1-k\delta'$ if holds that for every $i \in [k]$ we have $d\inparen{\x,S^{(i)}_\cx} \le \epsilon'$, in which case there are at least $k$ examples in $S$ with distance $\le \epsilon'$ to $\x$, so $S_\cx \in \cs$.

\noindent\textbf{Claim:}
$\prob{S \sim \cd^m}{\ca_{\kNN}(S)(\x) = f^*_\cd(\x) | S_\cx \in \cs} \ge \frac{1}{2} + \frac{\gamma(\cd)}{2} -\frac{\tau}{4}$

\noindent\textbf{Proof:} Fix some $S_\cx \in \cs$, and w.l.o.g. assume that $\kpi(\x,S_\cx) = \{\x_1, \x_2, \dots, \x_k\}$. Then,
\begin{align*}
&\prob{S' \sim \cd^m}{\ca_{\kNN}(S)(\x) = f^*_\cd(\x)\mid S'_\cx = S_\cx} \\
&= \prob{S' \sim \cd^m}{\sign\inparen{\sum_{i=1}^k y_i} = f^*_\cd(\x)\mid S'_\cx = S_\cx}
\end{align*}
Denote $p_i = \prob{S'\sim\cd^m}{y_i = f_\cd^*(\x)|S_\cx'=S_\cx} = \P_\cd[f^*_\cd(\x)|\x_i]$. Now, observe that:
\[
p_i \ge \P_\cd[f^*_\cd(\x)|\x] - \lambda d(\x,\x_i) \ge \P_\cd[f^*(\x)|\x]-\lambda\epsilon' \ge \frac{1}{2} + \frac{\gamma(\cd)}{2} - \lambda  \epsilon' \ge \frac{1}{2} + \frac{\gamma(\cd)}{2}-\frac{\tau}{4}
\]
where the first inequality uses the $\lambda$-Lipschitz property of $\cd$, and the third inequality is by definition of $\gamma(\cd)$. Now, from the Conodorcet Jury Theorem in \cite{ben2000nonasymptotic}, it holds that:
\[
\prob{S' \sim \cd^m}{\sign\inparen{\sum_{i=1}^k y_i} = f^*_\cd(\x)\mid S'_\cx = S_\cx} \ge \frac{1}{k} \sum_{i=1}^k p_i \ge \frac{1}{2} + \frac{\gamma(\cd)}{2}-\frac{\tau}{4}
\]
and the claim follows from the law of total probability.

\noindent\textbf{Claim:} For every $\x \in \cx'$ it holds that $\prob{S \sim \cd^m}{\ca_{\kNN}(S)(\x) = f^*_\cd(\x)} \ge \frac{1}{2} + \frac{\gamma(\cd)-\tau}{2}$.

\noindent\textbf{Proof:} Observe that, using the previous claims:
\begin{align*}
\prob{S \sim \cd^m}{\ca_{\kNN}(S)(\x) \ne f^*_\cd(\x)} &\le \prob{S \sim \cd^m}{\ca_{\kNN}(S)(\x) \ne f^*_\cd(\x) | S_\cx \in \cs} + \prob{S\sim\cd^m}{S \notin \cs} \\
&< \frac{1}{2} - \frac{\gamma(\cd)}{2} + \frac{\tau}{4} + k \delta' \le \frac{1}{2} - \frac{\gamma(\cd)-\tau}{2}
\end{align*}

By the previous claim, it follows that for all $\x \in \cx'$ we have $f^*_{\ca_{\kNN}(\cd^m)}(\x) = f^*_\cd(\x)$, and using the fact that $\prob{\x \sim \cd}{\x \notin \cx'} \le \delta' \le \epsilon$ the first condition for teacher holds. Since we also have $\prob{\x \sim \cd}{\x \notin \cx'} \le \delta' \le \delta$, by the previous claim we get that $\gamma_\delta(\ca_{\kNN}(\cd^m)) \ge \gamma(\cd) - \tau$, and the second condition in the definition of teacher holds.
\end{proof}

\begin{proof}{of Theorem \ref{thm:bounded_norm_relu}}.

Let $\epsilon >0$ and let $\eps'=\eps/4$.
We begin with the following claim:

\noindent\textbf{Claim}: There exist numbers $a < b$ such that $\prob{x \sim \cd}{x \le a} = \prob{x \sim \cd}{x \ge b} = \epsilon'$.

\noindent\textbf{Proof}: By assumption the function $F(a)=\P[x\le a]$ is continuous and $\lim_{a\to \infty}=1$, $\lim_{a\to -\infty}=0$ thus by the intermediate value theorem we have there exist $a,b$ such that $F(a)=\P[x\le a]=\eps'$ and $F(b)=\P[x\le b]=1-\eps'$.

Assume we sample $S \sim \cd^m$, and sort it s.t. $S = ((x_1,y_1), \dots, (x_m,y_m))$ where $x_1 < x_2 < \dots < x_m$.

\noindent\textbf{Claim}: Fix $\delta' > 0$, and assume that $m \ge \frac{\log(2/\delta')}{\epsilon'}$. Then, w.p. at least $1-\delta'$ over the choice of $S$, it holds that
\[
\prob{x \sim \cd}{x \notin [x_1,x_m]} \le 2 \epsilon'
\]

\noindent\textbf{Proof}: Let $a,b$ be the numbers guaranteed by the previous claim. Then, we have
\[
\prob{S \sim \cd^m}{x_1 > a} = \prob{S \sim \cd^m}{\forall (x,y) \in S~,~x > a} = (1-\epsilon')^m \le e^{-\epsilon'm} \le \frac{\delta'}{2}
\]
and similarly we get $\prob{S \sim \cd^m}{x_m < b}\le \frac{\delta'}{2}$. So, from the union bound, w.p. at least $1-\delta'$ it holds that $x_1 \le a$ and $x_m \ge b$. In this case, we have:
\[
\prob{x \sim \cd}{x \notin [x_1,x_m]} \le \prob{x \sim \cd}{x \notin (a,b)} = 2 \epsilon' = \eps/2
\]

\noindent\textbf{Claim}: Split $[a,b]$ to $\frac{b-a}{\delta}$ intervals of equal size of $\delta$ and denote the intervals by $A_i = [a+i\delta, a+(i+1)\delta)$. Then, letting $m\ge \frac{6(b-a)}{\eps\delta} \log (6(b-a)/\eps\delta)$ where $\delta = \eps/12\lambda$.

$$\P[\exists A_i \text{ s.t. } x \in A_i \text{ and } \forall x_j \in S, x_j \notin A_i]\le \eps/3.$$
\noindent\textbf{Proof}: Denote the above event by $B$. Now, let $p_i=\P[x\in A_i]$, by the union bound: 
\begin{align*}
\P(B) &\le \sum \P[x\in A_i, \forall x_j, x_j \notin A_i]\\
&= \sum_i p_i (1-p_i)^m \\
&\le \sum_i p_i e^{-p_i m} \\
&= \sum_{i:p_i <\frac{\delta}{b-a}\eps/6} p_i e^{-p_i m} + \sum_{i:p_i \ge\frac{\delta}{b-a}\eps/6} p_i e^{-p_i m} \\
&\le \eps/6 +  \sum_{i:p_i \ge\frac{\delta}{b-a}\eps/6} e^{-p_i m}
\end{align*}
Now, since we chose $m\ge \frac{6(b-a)}{\eps\delta}\log(6(b-a)/\eps\delta)$ we have that,
$$P(B)\le \eps/3.$$

\noindent\textbf{Claim}: When $m$ defined as above, we have that,
$$
\E_x \left[ \sum_{y} \abs{\P_{S \sim \cd^m}[\ca(S)(x) | x] - \P[y|x]}\bigg|x\in[a,b]\right] \le \eps/2
$$
\begin{proof}
Let $C$ be the event $x\in[a,b]$ intersected with $B^c=\Omega\setminus B$. Then, denote by  $x_i$ the nearest neighbor of $x$ in $S$ and assume WLOG $x\in [x_i, x_i+1]$. Conditioned on $C$, $x-x_i\le \delta$, thus, 
\begin{align*}
\abs{h_{\hat{\theta}}(x)-y_i} &= \abs{\frac{y_{i+1}-y_i}{x_{i+1}-x_i}\inparen{x-x_i}} \\ 
& \le \frac{|y_{i+1}-y_i|}{2} \le 1
\end{align*}
And consequently, the sign of $x$ will be $y_i$. Thus, (conditioning on $C$)

\begin{align*}
\abs{\P_{S \sim \cd^m}[\ca(S)(x) | x] - \P[y|x]} &\le \abs{\P_{S \sim \cd^m}[\ca(S)(x) | x]-\P[y|x_i]} + \abs{\P[y|x_i] - \P[y|x]} \\
& \le \lambda \delta \le \eps/12
\end{align*}
We thus can conclude that, 
\begin{align*}
\E_x &\left[\sum_{y} \abs{\P_{S \sim \cd^m}[\ca(S)(x) | x] - \P[y|x]}\bigg|x\in[a,b]\right] \\
& \le
\P(B) + \E_x\left[\sum_{y} \abs{\P_{S \sim \cd^m}[\ca(S)(x) | x] - \P[y|x]}\bigg|C\right] = \eps/2
\end{align*}

Now combining all of the above conclude the proof of the Theorem.
\end{proof}

\end{proof}

\begin{proof}{of Theorem \ref{thm:finite-classes}.}
First, we show that $\ca=\ERM_\ch$ is a teacher when $m=\frac{2k\log(2k/\eps)}{\eps}$. Let $S=\{(\x_i, y_i)\}_{i=1}^{m}\sim \cd^{m}$ be the sample set. Also,   let  $B = \{\x \in \cx\ | \ \P(\x_i=\x)\le \frac{\eps}{2k}\}$ and $G=\cx \setminus B$. We first show that for every $\x \in G$ we have $\P[\x\notin S]\le \frac{\eps}{2k}$.
\[
\P[\x\notin S | \x \in G]\le (1-\eps/2k)^m\le e^{-\log(2k/\eps)} = \eps/2k
\]

Now we can use the union bound to show that, $$\P[\exists \x_i \in G\setminus S]\le \eps/2$$
Using the union bound again, we can see that for a new example $\x'$: $\P[\x'\in B]\le \frac{\eps}{2}$.
Thus, $\P[\x'\in B \text{ or } \exists \x_i \in G\setminus S]\le \eps$. Now, for each $\x \in S\bigcap G$ the label $y(\x)$ given by $\ERM$ can be seen as a Condorcet Jury voting by the set of $\{y_i| \x_i = \x\}$. We can  use Theorem 1 from \citet{berend2005monotonicity} that shows that Condorcet Jury voting is monotone in the number of votes. Thus, $\P[f^*_\cd(\x') = f^*_{\ca(\cd^m)}(\x')] = 1$ using the aforementioned conditioning. Similarly, we have that  $\gamma_\eps(\ca(\cd^m)) \ge \gamma(\cd)$ (i.e., $\tau=0$). As we can condition as before and the CJT  monotonicity Theorem implies that  the margin can only increase (as the probability of the top label increases).
\end{proof}

\begin{lemma}
\label{lem:label_bound}
Let $y_1, \dots, y_n$ be some independent random variables with $y_i \in \{\pm 1\}$ s.t. $\P(y_i = 1) = p_i$, where either $p_1, \dots, p_n \in (1/2,1]$ or $p_1,\dots,p_n \in [0,1/2)$, and let $\gamma = \min_i \abs{2p_i-1}$. Denote $y^* = \sign(\sum_{i=1}^n y_i)$, and
let $\ell(\y) = 2 \cdot \sum_{i=1}^n \ind \{y_i \ne y^*\}$ and $\tilde{\ell}(\y,r) = n(1-r)$ for some $0 < r \le \frac{\gamma}{3}$.
Then, there exists some universal constant $c > 0$, s.t. for every $\delta \in (0,1)$, if $n\ge \frac{8\log(1/\delta)}{\gamma^2}$ w.p. at least $1-\delta$ we have $\ell(\y) < \tilde{\ell}(\y)$.
\end{lemma}


\begin{proof}
Let $S = \sum_{i=1}^n y_i$.
Observe that:
\[
\ell(\y) = 2 \sum_{i=1}^n\ind\{y_i \ne y^*\} = 2 \cdot \sum_{i=1}^n\inparen{\frac{1}{2}-\frac{y_iy^*}{2}} = n-y^* \sum_{i=1}^n y_i=n-\abs{S}
\]

 Also note that $\E[S] = \sum_{i=1}^n (2p_i-1)$ so $\abs{\E[S]} \ge n \gamma$. Now, from Hoeffding's inequality:
\[
\P\inparen{\abs{S-\E[S]} \ge \frac{n\gamma}{2}} \le 2\exp\inparen{-n\gamma^2/8} \le \delta
\]

So, w.p. at least $1-\delta$ we have:
\begin{align*}
    \ell(\y) = n-\abs{S} \le n-\abs{\E[S]} + \abs{S-\E[S]} \le n-\frac{n\gamma}{2} < n-rn = \tilde{\ell}(\y)
\end{align*}
where we use the fact that $r \le \frac{\gamma}{3} < \frac{\gamma}{2}$.
\end{proof}

\begin{proof}{of Theorem \ref{thm:lipschitz_class}.}

\paragraph{Claim.} The Bayes optimal classifier $f^*$ on $\cd$ is constant on each ball.
\paragraph{Proof.} Let $\x \in B(\bc_i, r)$ and let $y_1 =: f^*(\bc_i)$ be the $\arg\max\prob{\cd}{y|\bc_i}$. Using the margin condition on $\bc_i$ we know that $\P(y_1|\bc_i) > \P( y_2|\bc_i)+\gamma$ (here $y_2=-y_1$). Since $\x \in \cb(\bc_i)$ we know that $d(\x, \bc_i)<r$ and using the $\lambda$-Lipschitzness of the distribution we get that,
\[
\P(y_1|\x) \ge \P(y_1|\bc_i) - \lambda r > \P(y_2|\bc_i)+\gamma - \lambda r \ge \P(y_2|\x)+\gamma - 2\lambda r \ge \P(y_2|\x)
\]
So $f^*(\x)= f^*(\bc_i)$ thus $f^*$ on $B(\bc_i, r)$ is determined by $f^*(\bc_i)$ and therefore constant on the ball. \\
In a similar fashion, we proceed to show that with high probability a hypothesis output by $\ERM_\ch^{hinge}$ is constant on each ball with significant probability mass.
\paragraph{Claim.} Let $h \in \ch$ be some function that is not constant on $B(\bc_i,r)$. Then $\abs{\hat{h}(\x)} \le 2Lr$ for every $\x \in B(\bc_i,r)$.
\paragraph{Proof.} Fix $\x \in B(\bc_i,r)$ and let $\x' \in B(\bc_i,r)$ s.t. $\sign\hat{h}(\x) \ne \sign\hat{h}(\x')$.
Observe that $\abs{\hat{h}(\x)-\hat{h}(\x')} \le L\norm{\x-\x'} \le 2Lr$. So, if $\hat{h}(\x) > 0$ we get that
\[
\hat{h}(\x) \le \hat{h}(\x) - \hat{h}(\x') \le 2Lr
\]
otherwise if $\hat{h}(\x) \le 0$ we get that
\[
-\hat{h}(\x) \le \hat{h}(\x')-\hat{h}(\x) \le 2Lr
\]
\paragraph{Claim.} For each ball $B(\bc_i, r)$ with $\P[\x \in B(\bc_i, r)] \ge \eps/2k$, if $m \ge \frac{16k\log(2k/\eps)}{\gamma^2\eps}$ we have w.p. at least $1-\eps/2k$ that $\abs{S \cap B(\bc_i,r)} \ge n$ where $n=\frac{8\log(2k/\eps)}{\gamma^2}$.
\paragraph{Proof.} Let $S = \{(\x_i, y_i)\}_{i=1}^m$ and denote $\xi_i = \ind \{\x_i \in B(\bc_i,r)\}$, and notice that $\abs{S \cap B(\bc_i,r)} = \sum_{i=1}^m \xi_i$. It holds that: $\E\left[\sum_{i=1}^m \xi_i\right] \ge \frac{m\epsilon}{2k}$.
Note, similar to the argument in Theorem~\ref{thm:finite-classes}, if $m\ge\frac{2k\log(1/\delta)}{\eps}$ w.p. $1-\delta$ it holds that $\sum_{i=1}^m \xi_i\ge 1$. When $m\ge \frac{16k\log(2k/\eps)\log(16k\log(2k/\eps)/\eps\gamma^2)}{\eps \gamma^2}$ we can apply the same argument for each ``block" of size $\frac{2k\log(16k\log(2k/\eps)/\eps\gamma^2)}{\eps}$. That is, we are using $\delta=\frac{\eps \gamma^2}{16k\log(\frac{2k}{\eps})}$ and the number of blocks is $n=\frac{8\log(2k/\eps)}{\gamma^2}$ to get that with probability $1-\delta n = 1-\frac{\eps}{2k}$ it holds that $\sum_{i=1}^m \xi_i\ge \frac{8\log(2k/\eps)}{\gamma^2}$.
\paragraph{Claim.} For each ball $B(\bc_i, r)$ with $\P[\x \in B(\bc_i, r)] \ge \eps/2k$ the probability that $\ERM_\ch^{hinge}$ is constant on  $B(\bc_i, r)$ is at least $1-\eps/2k$.
\paragraph{Proof.} From the previous two claims it holds that $f^*$ is constant on $B(\bc_i,r)$ and that with probability $\ge 1-\eps/2k$ it holds that $|S\cap B(\bc_i,r)|\ge 8\log(2k/\eps)/\gamma^2$.  
Let $n = \abs{S \cap B(\bc_i,r)}$, and denote $(\x_1, y_1), \dots, (\x_n,y_n)$ the examples in $S \cap B(\bc_i,r)$. By definition of $\gamma(\cd)$ it holds that $\P_\cd[\tilde{y}_i=1|\bc_i] = p_i$ with $\abs{2p_i-1} \ge \gamma$. Let $y^* = \sign(\sum_{i=1}^n y_i)$ and let $h^* \in \ch$ be a hypothesis s.t. $h^*(\x) = y^*$. Let $h \in \ch$ be some function that is not constant on $B(\bc_i,r)$. Then:
\[
\sum_{i=1}^n \ell_{hinge}(h^*(\x_i), y_i) = 2 \sum_{i=1}^n \ind \{y_i \ne y^*\} = \ell(\y)
\]
Observe that from the previous claim we have $\abs{h(\x_i)} \le 2Lr < 1$ and therefore:
\[
\sum_{i=1}^n \ell_{hinge}(h(\x_i),y_i) \ge \sum_{i=1}^n 1-y_ih(\x_i) \ge \sum_{i=1}^n (1-\abs{h(\x_i)}) \ge n(1-2Lr) = \tilde{\ell}(\y,2Lr)
\]
Therefore, if $r \le \frac{\gamma}{3L}$, w.p. at least $1-\eps/2$ we have
$$\sum_{i=1}^n \ell_{hinge}(h^*(\x_i), y_i) \le \sum_{i=1}^n \ell_{hinge}(h(\x_i), y_i)$$

Thus, using the union bound we get that with with probability $>1-\eps$, $\ERM_{\ch}^{hinge}$ on each ball with probability mass $\ge\frac{\eps}{2k}$ will be constant. Now, since the Bayes is fixed on each ball, the output hypothesis could be seen as Condorocet Jury voting on each ball independently thus proving (same argument as Theorem-\ref{thm:finite-classes}) both condition 1 and 2.
\end{proof}

\section{Experimental Details}
\label{sec:exp-app}
In this section, we elaborate the exact details used in our experiments. In all experiments, we train ResNet-18 \citep{he2015} with batch size 128 and 0.0005
weight decay. On CIFAR-10 \citep{cifar10} we train for 50 epochs and for CIFAR-5m we train for 1 epoch using cos-annealing learning rate that starts from 0.05 for both datasets.   This optimization procedure achieves $\approx94\%$ accuracy on CIFAR-10 when train on clean data. However, we add 20\% fixed label noise. With label noise the model (without early stopping) has $81.3\%$ accuracy on the clean test set. For each experiment in the body we use (at-least) 10 random seeds. So for example, for the 10 random teachers experiment we train $100$ teacher models and chose 10 fixed teachers at random for each student seed.

\end{document}